\documentclass[11pt]{article}

\usepackage{acl}

\usepackage{times}
\usepackage{latexsym}
\usepackage[T1]{fontenc}
\usepackage[utf8]{inputenc}
\usepackage{microtype}
\usepackage{inconsolata}

\usepackage{graphicx}
\usepackage{subcaption}
\usepackage{multirow}
\usepackage{booktabs}
\usepackage[table]{xcolor}
\usepackage{amsmath}
\usepackage{amssymb}
\usepackage{mathtools}
\usepackage{amsthm}
\usepackage{placeins}
\usepackage[capitalize,noabbrev]{cleveref}
\usepackage{xspace}
\usepackage[most]{tcolorbox}
\usepackage{fvextra}
\usepackage{caption}
\usepackage{enumitem}
\usepackage{amsmath,amsfonts,bm}
\DeclareMathOperator*{\argmax}{arg\,max}

\newcommand{\medqbench}{\textbf{MedQ-Bench}\xspace}

\title{Can MLLMs Perceive and Describe Medical Image Quality Beyond Scores?}

\author{
 \textbf{Jiyao Liu\textsuperscript{1}$^{\ast}$},
 \textbf{Jinjie Wei\textsuperscript{3}$^{\ast}$},
 \textbf{Wanying Qu\textsuperscript{1}},
 \textbf{Chenglong Ma\textsuperscript{1}},
 \textbf{Junzhi Ning\textsuperscript{2}},
 \textbf{Yunheng Li\textsuperscript{1}},
 \\
 \textbf{Ying Chen\textsuperscript{2}},
 \textbf{Jiafan Zhang \textsuperscript{2}},
 \textbf{Pengcheng Chen\textsuperscript{1}},
 \textbf{Xin Gao\textsuperscript{1}},
 \textbf{Ming Hu\textsuperscript{2}},
 \textbf{Xin Wang\textsuperscript{2}},
\\
 \textbf{Shujian Gao\textsuperscript{1,2}},
 \textbf{Dingkang Yang\textsuperscript{1}},
 \textbf{Zhongying Deng\textsuperscript{4}},
 \textbf{Jin ye\textsuperscript{2}},
\\
 \textbf{Lihao Liu\textsuperscript{5}$^{\dagger}$},
 \textbf{Junjun He\textsuperscript{2}$^{\dagger}$},
 \textbf{Ningsheng Xu\textsuperscript{1}},
\\
\\
 \textsuperscript{1}Fudan University,
 \textsuperscript{2}Shanghai AI Lab,
 \textsuperscript{3}MBZUAI,
 \textsuperscript{4}Cambridge University,
 \textsuperscript{5}Amazon
\\
 \small{
   \textbf{Correspondence:} \href{mailto:lihaoliu@amazon.com}{lihaoliu@amazon.com}, \href{mailto:hejunjun@sjtu.edu.cn}{hejunjun@sjtu.edu.cn}
 }
}

\begin{document}
\maketitle
\renewcommand\thefootnote{}
\footnotetext{$^\ast$Equal contribution. $^\dagger$Corresponding authors.}
\renewcommand\thefootnote{\arabic{footnote}}

\begin{abstract}
Medical Image Quality Assessment (IQA) serves as a critical safety gate for clinical AI, yet existing approaches rely on scalar metrics that fail to capture the descriptive, expert-like evaluation process central to clinical practice. We introduce \medqbench, a comprehensive benchmark for language-based medical IQA evaluation with MLLMs, establishing a \textit{perception--description} paradigm across 5 imaging modalities and over 40 quality attributes. The benchmark comprises two complementary tasks: \textit{MedQ-Perception} probes low-level visual attribute discrimination via 2,600 expert-annotated perceptual queries; \textit{MedQ-Description} covers both no-reference and comparison description sub-tasks with 708 expert-annotated samples, evaluated through a four-dimensional judging protocol. Comprehensive evaluation of 23 state-of-the-art MLLMs reveals that current models exhibit only preliminary and unstable capabilities, and highlights the urgent need for targeted optimization of MLLMs in medical IQA. Project \& Dataset: 
\url{https://github.com/liujiyaoFDU/MedQ-Bench}

\end{abstract}

\begin{figure}[!ht]
  \begin{center}
    \centerline{\includegraphics[width=\columnwidth]{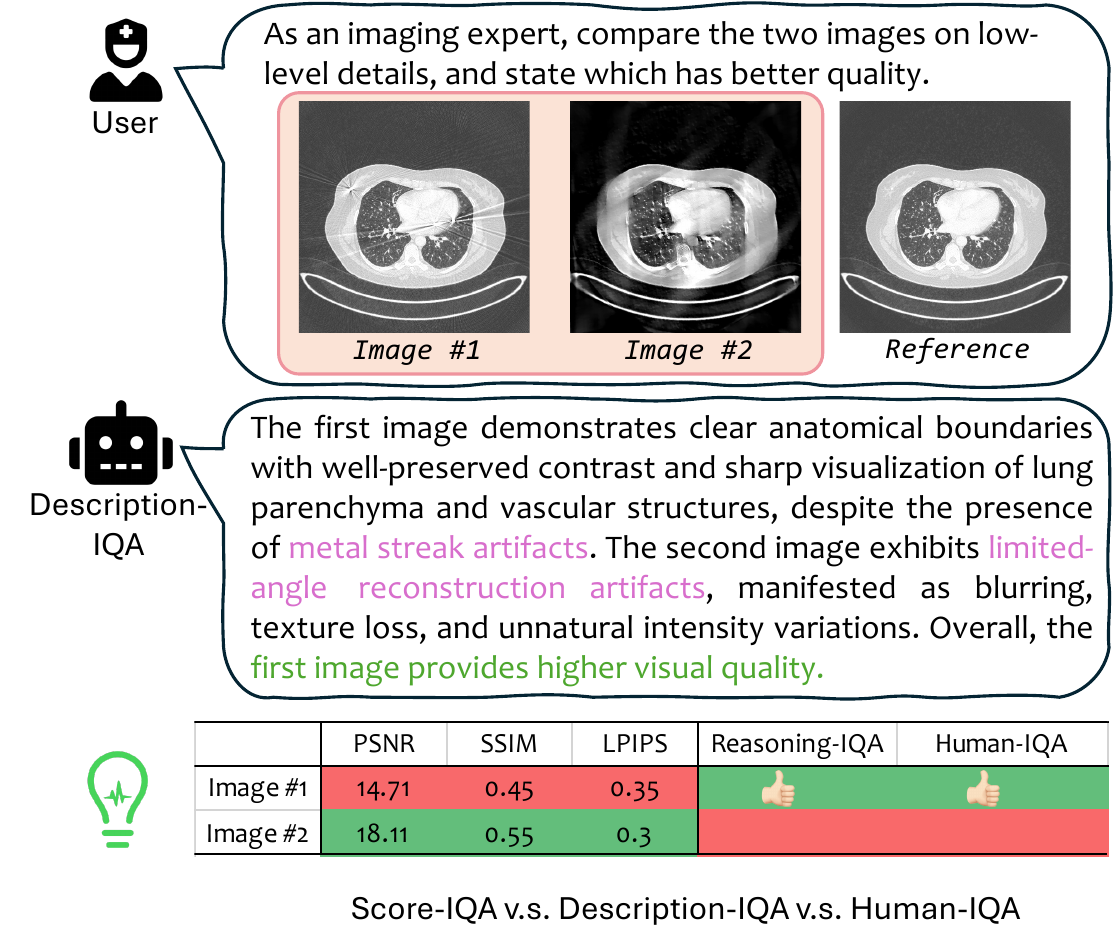}}
    \caption{
      Comparison of Description-IQA with score-based IQA paradigm. Unlike purely numerical scores, Description-IQA identifies distortion types and their relative impact, yielding results more consistent with Human-IQA.
    }
    \label{fig:compare_case}
  \end{center}
  \vspace{-12pt}
\end{figure}

\begin{figure*}[!t]
    \centering
    \includegraphics[width=0.9\textwidth]{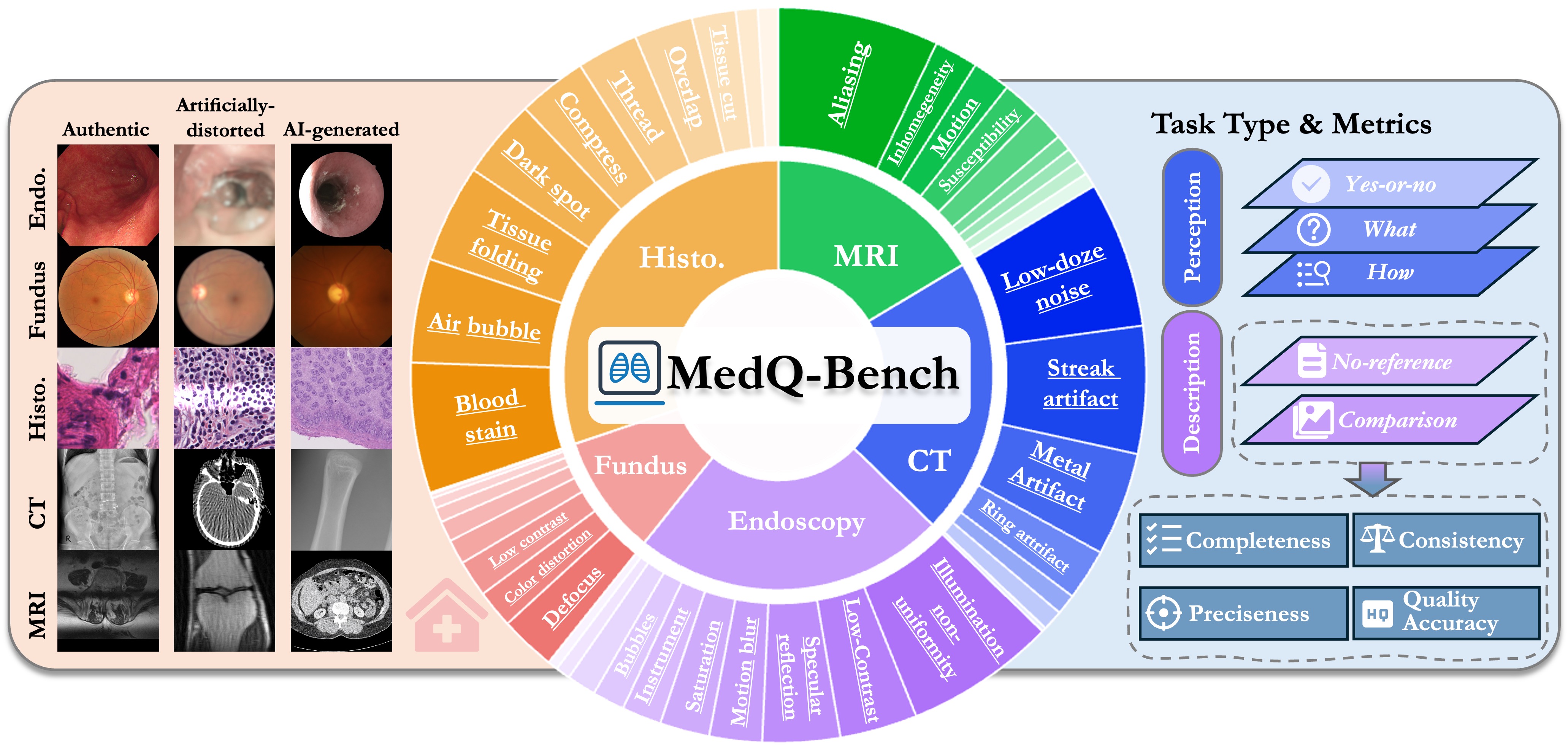}
    \caption{\medqbench overview. (1) Comprehensive coverage: 3,308 samples across 5 modalities with 40+ degradation types. (2) Multi-faceted evaluation: perception-description paradigm.}
    \label{fig:benchmark_overview}
\end{figure*}

\section{Introduction}

Medical Image Quality Assessment (IQA) examines whether medical images satisfy basic quality requirements for reliable downstream analysis and interpretation~\citep{tejani2024checklist}. In practice, degradations in low-level attributes such as contrast, noise, and artifacts affect image readability and result stability~\citep{liu2026medq}.

Current medical IQA approaches predominantly produce scalar scores using (1) no-reference methods~\citep{xun2025mediqa,herath2025systematic}, which infer perceptual quality of an image through statistical feature extraction without a reference, and (2) full-reference similarity metrics such as PSNR, SSIM~\citep{wang2004image}, and LPIPS~\citep{zhang2018unreasonable}. These methods provide standardized evaluation metrics and enable automated IQA. However, they exhibit the following fundamental deficiencies.
(1) \textit{Poor generalization}~\citep{herath2025systematic}. Medical image quality is influenced by complex and heterogeneous factors, including modality-specific degradations 
across magnetic resonance imaging (MRI), computed tomography (CT), fundus photography (Fundus), histopathology, and endoscopy. Yet, existing methods typically rely on simple regression models~\citep{su2023deep} or handcrafted statistical indices~\citep{dohmen2024five}, which generalize poorly to unseen distortions, new modalities, or different imaging protocols.
(2) \textit{Lack of human-like description process for result interpretation}. Most methods produce scalar IQA scores, which do not fully reflect the causes of image quality degradation and may be unreasonable in certain cases. As illustrated in Figure~\ref{fig:compare_case}, clinicians identify specific degradations and assess their clinical impact before making quality judgments. Traditional score-based metrics may contradict human judgment by favoring superficially smoother images. Such description requires understanding the quality factors, which current paradigm approaches cannot effectively capture.

Recent advances in multimodal large language models (MLLMs) have shown promising capabilities in medical visual description tasks~\citep{ liu2023visual,dai2023instructblip,sellergren2025medgemma,su2025gmai}. Theoretically, MLLMs could potentially address existing IQA challenges by decomposing quality assessment into interconnected subtasks: degradation identification, severity quantification, impact analysis, and comparative description. Unlike traditional approaches that yield scores, MLLM-based assessment can provide explicit chains of thought~\citep{wu2023q,you2024depicting}, offering interpretable and clinically meaningful evaluations.
However, critical questions remain unanswered about MLLMs' actual capabilities in medical IQA: Can they truly generalize to the fine-grained, diverse, and complex quality factors spanning different imaging modalities? Do they possess genuine description abilities to understand the clinical significance of various degradations? Existing MLLM evaluation frameworks focus mainly on natural images~\citep{wu2024q} or high-level medical semantics~\citep{ye2024gmai}, lacking systematic benchmarks that assess quality-related perceptual and description skills across diverse medical modalities. This absence of specialized benchmarks has been a major barrier to developing and validating effective frameworks.

\begin{figure*}[!t]
    \centering
    \includegraphics[width=1.0\textwidth]{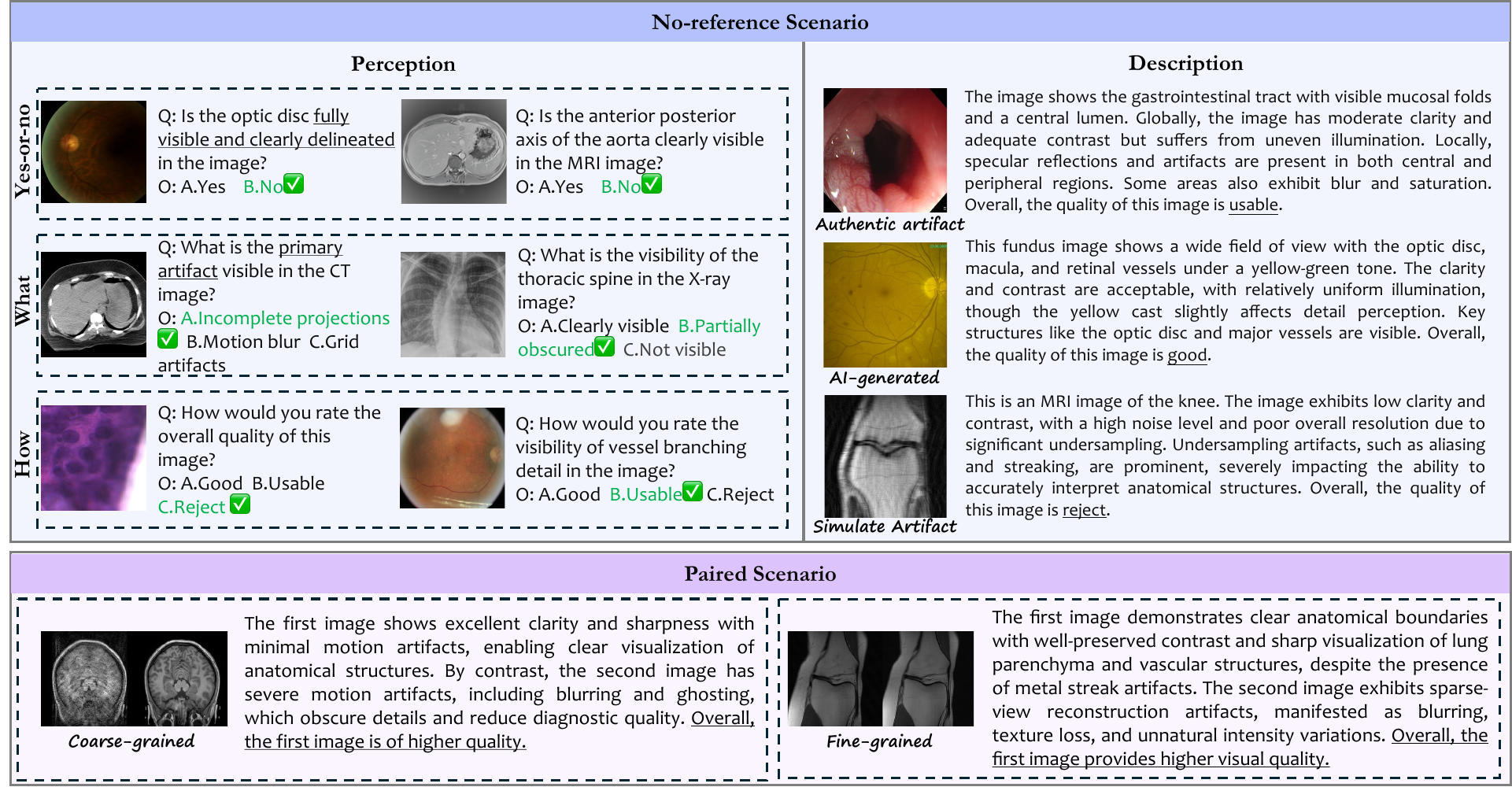}
    \caption{Examples of question types in MedQ-Bench, covering MCQA perception tasks, open-ended description, and pair comparison.}
    \label{fig:question_examples}
\end{figure*}

To bridge the gap between existing medical IQA methods, we propose a novel \emph{perception--description paradigm}. This paradigm mirrors clinicians' cognitive workflow: first perceiving quality-related attributes in images, assessing their severity, and evaluating their potential impact on clinical diagnosis, and then making overall quality judgments through logical description. Building on this paradigm, we introduce \medqbench, the first comprehensive benchmark that systematically evaluates the medical IQA capabilities of MLLMs (\Cref{fig:benchmark_overview}). Our primary contributions are as follows:

\begin{itemize}[leftmargin=*, itemindent=0pt]
\item \medqbench proposes a comprehensive methodology to assess MLLMs' perceptual and description abilities on medical images, supporting both no-reference and full-reference paradigms across various discrimination tasks.
\item Covering 5 imaging modalities and 40+ quality attributes, the dataset includes real, degraded, and AI-generated images across a wide range of clinical and controlled scenarios.
\item We introduce a four-axis scoring protocol for description evaluation and validate its reliability through comparison with radiologist assessments.
\item  Evaluation of both general-purpose and specialized MLLMs reveals significant gaps in Medical IQA, highlighting the need for improvements.
\end{itemize}

\section{Constructing the MedQ-Bench}

This section presents MedQ-Bench construction and task design: one MCQA perception task and two description tasks, as illustrated in \Cref{fig:question_examples}.

\subsection{Benchmark Scope and Modalities}

Image quality is a fundamental factor in medical imaging analysis, yet existing evaluation methods still rely primarily on score-based metrics and do not fully characterize quality perception and description capabilities. MedQ-Bench is specifically designed to systematically evaluate the visual quality perception and description capabilities of multimodal large language models (MLLMs) within the medical imaging domain. Let $\mathcal{M} = \{M_1, M_2, \ldots, M_m\}$ represent the set of medical imaging modalities, where each modality $M_i$ is associated with a distinct set of quality attributes $\mathcal{A}_i = \{a_{i,1}, a_{i,2}, \ldots, a_{i,k}\}$. The quality assessment task can be formulated as learning a mapping function $f: \mathcal{I} \times \mathcal{Q} \rightarrow \mathcal{R}$ that takes an image $I \in \mathcal{I}$ and question $q \in \mathcal{Q}$ as input and produces a response $r \in \mathcal{R}$.

To capture the diversity and complexity of real-world clinical imaging, MedQ-Bench encompasses five representative modalities: Magnetic Resonance Imaging (MRI), Computed Tomography (CT), endoscopy, histopathology imaging, and fundus photography. Let $\mathcal{D}_i = \{d_{i,1}, d_{i,2}, \ldots, d_{i,n}\}$ denote the set of degradation types specific to modality $M_i$. Each modality exhibits distinct degradation characteristics due to its physical acquisition principles, where degradations can be modeled as transformations $T_d: \mathcal{I} \rightarrow \mathcal{I}'$ that modify the original image $I$ based on degradation type $d \in \mathcal{D}_i$. For instance, MRI is particularly susceptible to motion and magnetic susceptibility artifacts, and CT is prone to low-dose noise and metal-induced streak artifacts. This multi-modality design ensures that the benchmark reflects the broad spectrum of perceptual challenges encountered in practice.

For each modality, MedQ-Bench incorporates images from three complementary sources: authentic clinical images containing naturally occurring artifacts; synthetically degraded images that replicate modality-specific distortions in a controlled manner; and AI-generated or reconstructed images produced by enhancement, translation, or reconstruction models, which may introduce hallucinations or subtle structural inconsistencies. This tri-source strategy enables the benchmark to cover both naturally occurring degradations and algorithm-induced artifacts, ensuring a balanced evaluation of MLLM robustness across real-world and algorithmic distortion scenarios.

\subsection{Benchmark on IQA Perception Ability}

Before evaluating sophisticated description capabilities, it is essential to establish whether MLLMs possess fundamental perceptual abilities to recognize basic quality attributes. 

\subsubsection{Question Types}

The perception-focused MCQA setting evaluates direct visual perception using single-image prompts, without requiring domain-specific diagnostic description. These tasks represent the most basic level of quality assessment capability, asking models to simply identify ``what they see'' rather than explain ``why they see it.'' For each image, three canonical subtypes of questions are included:
\textbf{\textit{(1) Yes-or-No}}: Binary classification tasks where $\mathcal{R}_\text{YN} = \{0, 1\}$ and the model predicts $\hat{y} = \argmax_{y \in \{0,1\}} P(y \mid I, q)$. Examples include ``Is this image clear?'' or ``Does this image contain artifacts?'' 
\textbf{\textit{(2) What}}: Multi-class identification tasks where $\mathcal{R}_\text{What} = \{c_1, c_2, \ldots, c_K\}$ represents $K$ possible degradation types, and the model selects $\hat{c} = \argmax_{c \in \mathcal{R}_\text{What}} P(c\mid I, q)$. These tasks ask models to identify specific types of artifacts or degradations present in the image.
\textbf{\textit{(3) How}}: Severity assessment tasks where $\mathcal{R}_\text{How} = \{s_1, s_2, \ldots, s_L\}$ represents $L$ severity levels, and the model predicts $\hat{s} = \argmax_{s \in \mathcal{R}_\text{How}} P(s\mid I, q)$. These tasks evaluate the model's ability to assess the degree or intensity of observed quality issues.

\subsubsection{Quadrants for Low-Level Visual Concerns}

\textbf{Axis 1: No Degradation vs Degradation Severity Levels.} The primary axis differentiates medical images based on their quality degradation status: 1) \textbf{\textit{No Degradation}} refers to medical images that maintain optimal quality standards without artifacts or distortions, and 2) degradation with Severity Levels encompasses images with varying degrees of quality issues, further subdivided into \textbf{\textit{mild Degradation}} and \textbf{\textit{severe Degradation}}. 

\textbf{Axis 2: General Medical Questions vs Modality-specific Questions.} Quality perception in medical imaging intertwines with modality-specific technical characteristics. For instance, motion artifacts manifest differently in MRI versus CT scans. We curate \textbf{\textit{modality-specific questions}} that require understanding unique technical characteristics of specific imaging modalities (e.g., ``Does this MRI show susceptibility artifacts?''), while \textbf{\textit{general medical questions}} focus on universal quality concepts applicable across modalities (e.g., ``Is this image clear?''). This distinction evaluates both fundamental quality perception and specialized modality knowledge.

\subsection{Benchmark on IQA Description Ability}

\subsubsection{No-reference Description Tasks}

Building on MCQA ability evaluation, we further construct description tasks to assess whether a model can autonomously describe and explain quality-related observations in natural language. These tasks require generating comprehensive responses $w_{1:T} = \{w_1, w_2, \ldots, w_T\}$ that systematically detail multiple aspects of image quality assessment: \textit{(1)} modality and anatomical region identification; \textit{(2)} specific quality degradation characterization including type and severity; \textit{(3)} technical attribution of underlying causes; \textit{(4)} assessment of diagnostic impact and clinical implications; and \textit{(5)} definitive quality judgment with good/usable/reject recommendation. The description tasks evaluate whether models can perform structured quality analysis that mirrors expert clinical assessment, moving beyond simple classification to demonstrate understanding of the relationship between technical image properties and degradation mechanisms.

\subsubsection{Comparison Description Tasks}

Many clinical workflows require comparative quality assessment between two versions of the same study, such as ``original vs. reconstructed'' or outputs from competing reconstruction algorithms. For image pairs $(I_A, I_B)$, the comparative task seeks to determine preference $P(I_A \succ I_B)$ based on overall quality assessment. Models must identify which image exhibits higher diagnostic quality and provide detailed explanations for their judgment, such as explaining why one reconstruction algorithm preserves anatomical detail better than another.

\begin{table*}[!th]
    \centering
    \small
    \resizebox{\textwidth}{!}{
    \begin{tabular}{lccccccccc}
    \toprule
    \textbf{Sub-categories} & \multicolumn{4}{c}{\textbf{Perception}} & \multicolumn{5}{c}{\textbf{Description}} \\
    \cmidrule(lr){2-5} \cmidrule(lr){6-10}
    \textbf{Model (variant)} & \textbf{Yes-or-No$\uparrow$} & \textbf{What$\uparrow$} & \textbf{How$\uparrow$} & \textbf{Overall$\uparrow$} & \textbf{Comp.$\uparrow$} & \textbf{Prec.$\uparrow$} & \textbf{Cons.$\uparrow$} & \textbf{Qual.$\uparrow$} & \textbf{Overall$\uparrow$} \\
    \midrule
    \textit{random guess} & 50.00\% & 28.48\% & 33.30\% & 37.94\% &  &  &  &  &  \\
    \rowcolor{gray!15} \textit{Non-experts} & 67.50\% & 57.50\% & 57.50\% & 62.50\% & - & - & - & - & - \\
    \rowcolor{gray!15} \textit{Human experts} & 88.50\% & 77.50\% & 77.50\% & 82.50\% & 1.524 & 1.483 & 1.887 & 1.623 & 6.502 \\
    \midrule
    \noalign{\vskip -\belowrulesep}
    \multicolumn{10}{c}{\cellcolor{green!8}\textbf{\textit{Open-Source General Models}}} \\
    Qwen2.5-VL-Instruct (7B)    & \cellcolor{red!1}57.89\% & \cellcolor{red!16}48.45\% & \cellcolor{red!7}54.40\% & \cellcolor{red!6}54.71\%      & \cellcolor{red!24}0.715 & \cellcolor{red!27}0.670 & \cellcolor{green!43}1.855 & 1.127 & \cellcolor{red!2}4.367 \\
    Qwen2.5-VL-Instruct (32B)   & \cellcolor{green!19}67.38\% & \cellcolor{red!25}43.02\% & \underline{58.69\%} & \cellcolor{green!1}59.31\%      & \cellcolor{red!3}1.077 & \cellcolor{red!11}0.928 & \cellcolor{green!50}\textbf{1.977} & \cellcolor{green!10}1.290 & \cellcolor{green!33}\underline{5.272} \\
    InternVL3 (8B)              & \cellcolor{green!28}72.04\% & \cellcolor{red!18}47.67\% & \cellcolor{red!9}52.97\% & \cellcolor{green!3}60.08\%      & \cellcolor{red!11}0.928 & \cellcolor{red!14}0.878 & \cellcolor{green!43}1.858 & \cellcolor{green!12}1.317 & \cellcolor{green!21}4.983 \\
    InternVL3 (38B)             & \cellcolor{green!24}69.71\% & \cellcolor{red!2}\underline{57.36\%} & \cellcolor{red!9}52.97\% & \cellcolor{green!5}61.00\%      & \cellcolor{red!9}0.964 & \cellcolor{red!18}0.824 & \cellcolor{green!43}\underline{1.860} & \cellcolor{green!12}1.317 & \cellcolor{green!20}4.965 \\
    Qwen3-VL-235B               & \cellcolor{red!5}55.44\% & \cellcolor{green!20}67.82\% & \cellcolor{green!16}66.27\% & \cellcolor{green!8}62.45\% & \cellcolor{red!10}0.946 & \cellcolor{red!24}0.720 & \cellcolor{green!30}1.629 & \cellcolor{green!21}1.471 & \cellcolor{green!12}4.765 \\
    Qwen2.5-VL-Instruct (72B)   & \cellcolor{green!43}\underline{78.67\%} & \cellcolor{red!26}42.25\% & \cellcolor{red!4}56.44\% & \cellcolor{green!10}\underline{63.14\%}      & \cellcolor{red!13}0.905 & \cellcolor{red!15}0.860 & \cellcolor{green!45}1.896 & \cellcolor{green!12}1.321 & \cellcolor{green!21}4.982 \\
    Qwen3.5 (397B)                & \cellcolor{green!15}65.78\% & \cellcolor{green!18}67.17\% & \cellcolor{green!15}65.46\% & \cellcolor{green!17}66.67\% & \cellcolor{red!12}0.923 & \cellcolor{red!23}0.733 & \cellcolor{green!16}1.396 & \cellcolor{green!20}1.466 & \cellcolor{green!2}4.518 \\
    \midrule
    \noalign{\vskip -\belowrulesep}
    \multicolumn{10}{c}{\cellcolor{yellow!20}\textbf{\textit{Quality-Oriented Models}}} \\
    Q-Instruct                & \cellcolor{green!42}78.32\% & \cellcolor{red!36}36.43\% & \cellcolor{red!20}46.22\% & \cellcolor{red!1}58.01\% & \cellcolor{red!43}0.398 & \cellcolor{red!47}0.330 & \cellcolor{green!22}1.498 & \cellcolor{green!3}1.167 & \cellcolor{red!20}3.394 \\
    Q-Insight                 & \cellcolor{green!14}65.05\% & \cellcolor{red!21}45.74\% & \cellcolor{green!6}61.35\% & \cellcolor{green!3}59.85\% & \cellcolor{red!31}0.597 & \cellcolor{red!37}0.502 & \cellcolor{red!47}0.340 & \cellcolor{red!8}0.977 & \cellcolor{red!38}2.416 \\
    VisualQuality-R1          & \cellcolor{green!6}61.65\% & \cellcolor{red!20}46.12\% & \cellcolor{red!6}55.01\% & \cellcolor{red!4}56.09\% & \cellcolor{red!22}0.747 & \cellcolor{red!27}0.665 & \cellcolor{green!39}1.783 & \cellcolor{green!3}1.172 & \cellcolor{red!2}4.367 \\
    \midrule
    \noalign{\vskip -\belowrulesep}
    \multicolumn{10}{c}{\cellcolor{blue!8}\textbf{\textit{Medical-Specialized Models}}} \\
    BiMediX2 (8B)                & \cellcolor{red!22}44.98\% & \cellcolor{red!50}27.52\% & \cellcolor{red!50}27.81\% & \cellcolor{red!38}35.10\%      & \cellcolor{red!44}0.376 & \cellcolor{red!43}0.394 & \cellcolor{red!50}0.281 & \cellcolor{red!27}0.670 & \cellcolor{red!50}1.721 \\
    Lingshu (32B)                & \cellcolor{red!13}50.36\% & \cellcolor{red!13}50.39\% & \cellcolor{red!11}51.74\% & \cellcolor{red!13}50.88\%      & \cellcolor{red!30}0.624 & \cellcolor{red!25}0.697 & \cellcolor{green!47}1.932 & \cellcolor{red!4}1.059 & \cellcolor{red!3}4.312 \\
    MedGemma (27B)               & \cellcolor{green!18}67.03\% & \cellcolor{red!17}48.06\% & \cellcolor{red!13}50.72\% & \cellcolor{red!2}57.16\%      & \cellcolor{red!23}0.742 & \cellcolor{red!39}0.471 & \cellcolor{green!27}1.579 & \cellcolor{green!8}1.262 & \cellcolor{red!8}4.054 \\
    \midrule
    \noalign{\vskip -\belowrulesep}
    \multicolumn{10}{c}{\cellcolor{orange!15}\textbf{\textit{Commercial MLLMs}}} \\
    Mistral-Medium-3          & \cellcolor{green!16}65.95\% & \cellcolor{red!16}48.84\% & \cellcolor{red!9}52.97\% & \cellcolor{red!2}57.70\%      & \cellcolor{red!12}0.923 & \cellcolor{red!23}0.729 & \cellcolor{green!26}1.566 & \cellcolor{green!13}1.339 & \cellcolor{green!3}4.557 \\
    Claude-4-Sonnet           & \cellcolor{green!27}71.51\% & \cellcolor{red!20}46.51\% & \cellcolor{red!7}54.60\% & \cellcolor{green!3}60.23\%      & \cellcolor{red!23}0.742 & \cellcolor{red!29}0.633 & \cellcolor{green!38}1.778 & \cellcolor{green!15}1.376 & \cellcolor{green!2}4.529 \\
    Gemini-2.5-Pro            & \cellcolor{green!35}75.13\% & \cellcolor{red!6}\underline{55.02\%} & \cellcolor{red!13}50.54\% & \cellcolor{green!7}61.88\%      & \cellcolor{red!14}0.878 & \cellcolor{red!14}\underline{0.891} & \cellcolor{green!33}1.688 & \cellcolor{green!26}\textbf{1.561} & \cellcolor{green!23}5.018 \\
    Claude-Sonnet-4.6         & \cellcolor{green!17}66.67\% & \cellcolor{green!3}59.83\% & 58.63\% & \cellcolor{green!9}62.99\% & 1.109 & \cellcolor{red!15}0.873 & \cellcolor{green!22}1.498 & \cellcolor{green!22}1.489 & \cellcolor{green!21}4.968 \\
    Gemini-3.1-Pro            & \cellcolor{green!16}66.13\% & 58.32\% & \cellcolor{green!12}64.26\% & \cellcolor{green!9}63.07\% & \cellcolor{red!21}0.765 & \cellcolor{red!21}0.760 & \cellcolor{green!31}1.647 & \cellcolor{green!26}1.557 & \cellcolor{green!11}4.729 \\
    Grok-4                    & \cellcolor{green!31}73.30\% & \cellcolor{red!16}48.84\% & \cellcolor{green!1}\textbf{59.10\%} & \cellcolor{green!10}\underline{63.14\%}      & \cellcolor{red!8}\underline{0.982} & \cellcolor{red!16}0.846 & \cellcolor{green!40}1.801 & \cellcolor{green!16}1.389 & \cellcolor{green!22}5.017 \\
    Kimi-K2.5                 & \cellcolor{green!12}64.35\% & \cellcolor{green!12}64.36\% & \cellcolor{green!9}63.05\% & \cellcolor{green!12}64.06\% & \cellcolor{red!16}0.846 & \cellcolor{red!30}0.615 & \cellcolor{green!11}1.312 & \cellcolor{green!13}1.348 & \cellcolor{red!6}4.122 \\
    GPT-4o                    & \cellcolor{green!42}\underline{78.48\%} & \cellcolor{red!15}49.64\% & \cellcolor{red!2}57.32\% & \cellcolor{green!13}\underline{64.79\%}      & \cellcolor{red!7}\underline{1.009} & \cellcolor{red!5}\underline{1.027} & \cellcolor{green!44}\underline{1.878} & \cellcolor{green!17}\underline{1.407} & \cellcolor{green!35}\underline{5.321} \\
    GPT-5.4                   & \cellcolor{green!12}64.35\% & \cellcolor{green!20}68.03\% & \cellcolor{green!20}67.87\% & \cellcolor{green!17}66.59\% & \cellcolor{red!8}0.977 & \cellcolor{red!20}0.783 & \cellcolor{green!29}1.620 & \cellcolor{green!20}1.453 & \cellcolor{green!15}4.833 \\
    GPT-5                     & \cellcolor{green!50}\textbf{82.26\%} & \cellcolor{green!4}\textbf{60.47\%} & \underline{58.28\%} & \cellcolor{green!22}\textbf{68.97\%} & \cellcolor{green!5}\textbf{1.195} & \textbf{1.118} & \cellcolor{green!42}1.837 & \cellcolor{green!24}\underline{1.529} & \cellcolor{green!50}\textbf{5.679} \\
    \bottomrule
    \end{tabular}
    }
    \caption{Performance of different models on the MCQA perception and description tasks.}
    \label{tab:mcqa_results}
    \vspace{-5pt}
    \end{table*}

Comparative tasks are further categorized by the perceptual gap between images. \textit{1) Coarse-grained} comparisons involve clearly visible quality differences, making them straightforward for both humans and models. \textit{2) Fine-grained} comparisons involve subtle differences in noise patterns, contrast, or structure fidelity, requiring sensitivity to nuanced quality cues that may only be apparent upon careful inspection. This design enables separate evaluation of basic discrimination ability and advanced perceptual subtlety that approaches expert-level assessment sensitivity

\subsubsection{Evaluation Metrics}

\paragraph{Multi-dimensional judging protocol}
Description tasks are open-ended and inherently subjective, so we use an LLM-based automatic judging protocol. Following prior LLM-as-a-judge practice~\citep{liu-etal-2023-g}, we adopt GPT-4o as the judge model. Under a unified prompt and rubric, each response is scored on four dimensions with a three-level scale (0, 1, 2): \textbf{(1) Completeness}, whether the response covers key low-level visual information in the reference answer; \textbf{(2) Preciseness}, whether the response is consistent with the reference answer without clear factual or semantic contradictions; \textbf{(3) Consistency}, whether the description reasoning is logically aligned with the final quality conclusion; and \textbf{(4) Quality Accuracy}, whether the final quality conclusion matches the reference answer, or correctly identifies the higher-quality image in comparison tasks. The four dimension scores are aggregated as the overall result. To ensure reliability, we further validate human-judge alignment, with details reported in Section~\ref{sec:validation_human_alignment}.

\subsection{Dataset Construction and Annotation}

Medical imaging experts designed structured seed templates defining the semantic intent and question structure for each question type. These seeds were expanded using GPT-4o as a template instantiation tool under strict semantic constraints, with each generated question reviewed by three independent experts to eliminate model-specific linguistic biases. All question-answer pairs underwent multi-round expert validation: \textit{(1)} independent annotation by three specialists following standardized protocols and modality-specific checklists; \textit{(2)} cross-validation sessions where annotators identified divergent cases and conducted structured discussions; and \textit{(3)} consensus resolution with majority voting for conflicts. Complete details are provided in Appendix~\ref{sec:appendix_construction}.

\section{Results}

\begin{figure*}[!th]
  \centering
  \begin{minipage}[b]{0.52\textwidth}
      \centering
      \includegraphics[width=\textwidth]{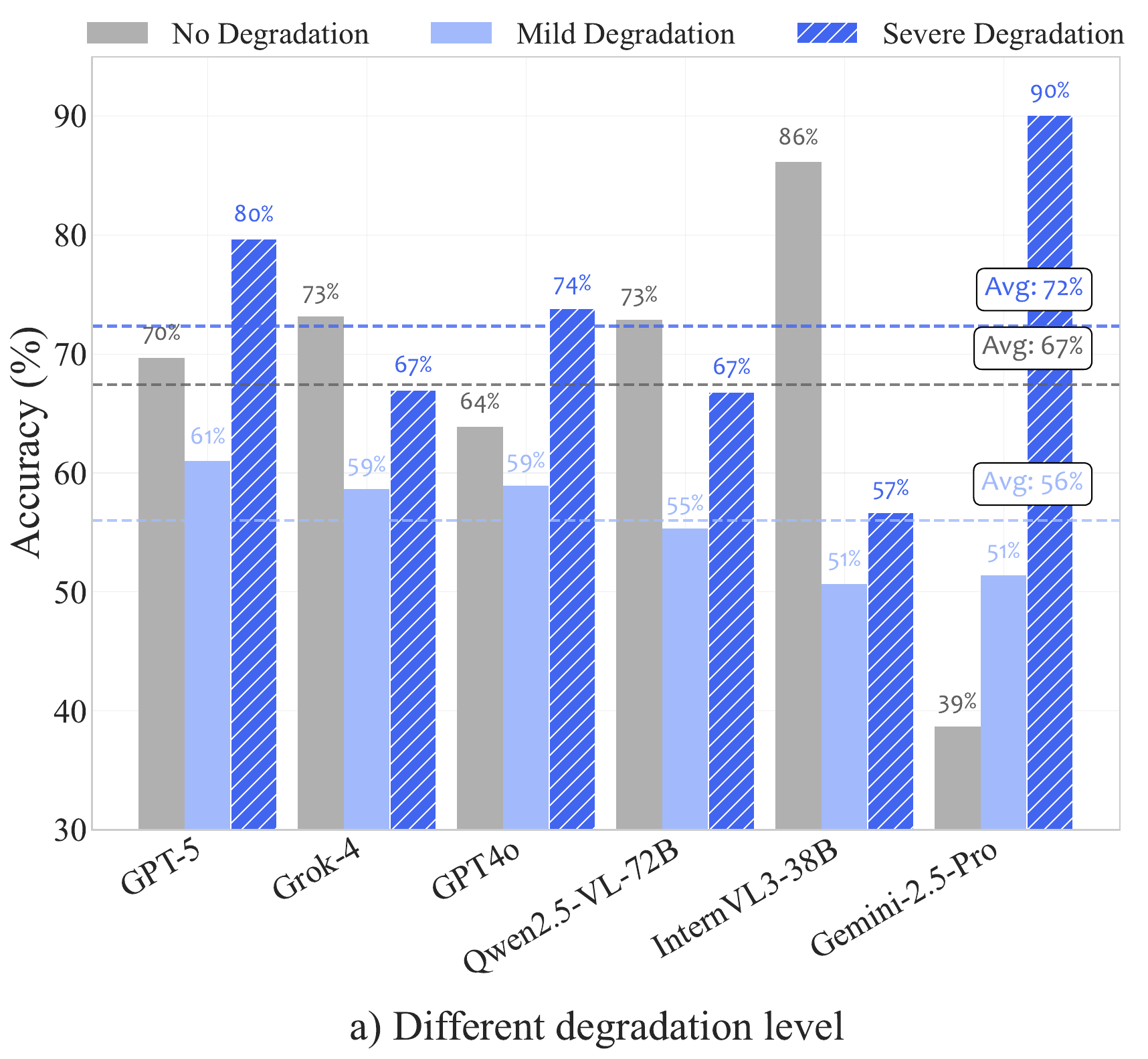}
  \end{minipage}
  \hfill
  \begin{minipage}[b]{0.40\textwidth}
      \centering
      \includegraphics[width=\textwidth]{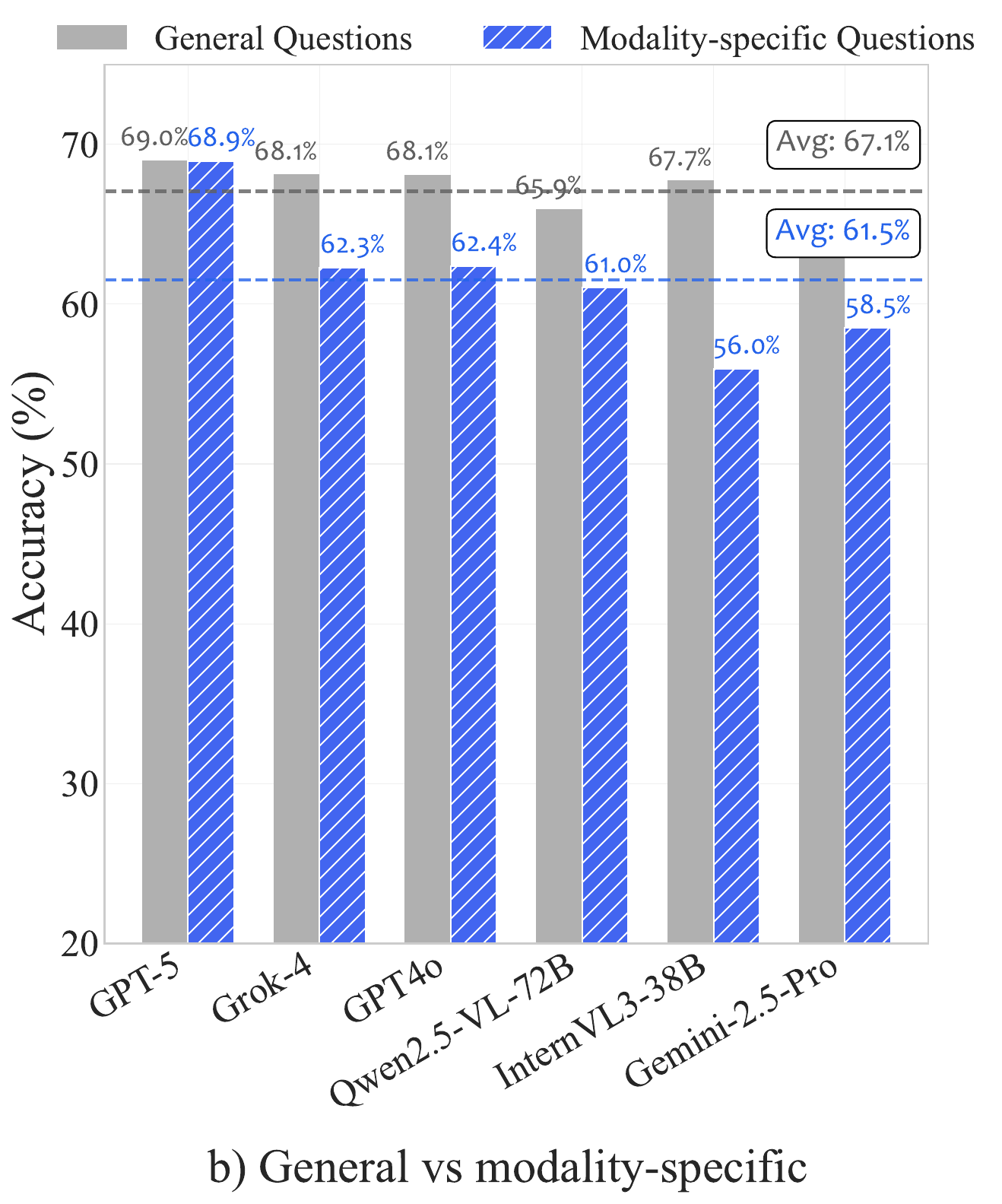}
  \end{minipage}
  \caption{Performance analysis of MLLMs across different evaluation dimensions: (a) Different degradation level performance, (b) general vs modality-specific question.}
  \label{fig:performance_analysis}
  
\end{figure*}

To investigate MLLMs' image quality perception ability, we present a comprehensive evaluation of MedQ-Bench across 23 up-to-date models under zero-shot settings. We evaluate these models across four categories: open-source MLLMs (Qwen2.5-VL-Instruct variants~\citep{wang2024qwen2}, InternVL3 models~\citep{chen2024internvl}, and Qwen3-family models), quality-oriented models (Q-Instruct~\citep{wu2024q}, Q-Insight~\citep{li2026q}, and VisualQuality-R1~\citep{wu2026visualquality}), medical-specialized MLLMs (BiMediX2~\citep{mullappilly2024bimedix2}, Lingshu~\citep{xu2025lingshu}, MedGemma~\citep{sellergren2025medgemma}), and general-purpose commercial systems (GPT-5 and GPT-5.4~\citep{openai2025gpt5}, GPT-4o~\citep{achiam2023gpt}, Gemini-2.5-Pro and Gemini-3.1-Pro~\citep{comanici2025gemini}, Grok-4, Claude-4-Sonnet and Claude-Sonnet-4.6~\citep{anthropic2025claude4systemcard}, Mistral-Medium-3~\citep{jiang2023mistral7b}, and Kimi-K2.5~\citep{team2026kimi}). 

\subsection{Findings on Perception}

\textbf{Conclusion 1. Clear performance hierarchy emerges across model categories:} Our analysis reveals that most MLLMs perform above random guessing across all sub-tasks, indicating promising potential for domain generalization. The results demonstrate a clear performance hierarchy: closed-source frontier models achieve the highest scores, with GPT-5 leading at 68.97\%. Among open-source models, Qwen3.5 (397B) achieves the best performance, outperforming most commercial models, while \underline{\textit{the medical models and quality-oriented models}} \underline{\textit{underperform expectations}}. 
More details are in \Cref{sec:why_medical_specialized_models_underperform}.
 
\textbf{Insufficiency 1. Substantial human-AI performance gap remains:} Another key finding emerges from our comparison with human performance, where we include both \textbf{human experts} (medical imaging technicians and medical imaging PhDs) and \textbf{non-experts} as reference points. The best AI model (GPT-5) significantly underperforms human experts, yet outperforms non-experts by 6.47\%. Given that these models have not undergone specialized training for medical image quality assessment, this suggests substantial potential for improvement in these MLLMs through further fine-tuning.

\textbf{Insufficiency 2. The MLLMs are not robust among different perceptual types:} Task-specific analysis reveals distinct patterns across different evaluation dimensions. Performance analysis across different degradation levels (Figure~\ref{fig:performance_analysis} (a)) demonstrates that mild degradation represents the most challenging scenario, with average accuracy dropping to 56\% compared to 72\% for no degradation and 67\% for severe degradation. This indicates that subtle quality issues are harder to identify than obvious artifacts. Top-performing models like GPT-5 demonstrate a degree of consistency in performance across degradation levels. We further investigate the difference between general and modality-specific medical questions. As shown in Figure~\ref{fig:performance_analysis}(b), most models perform better on general questions than on modality-specific tasks, whereas GPT-5 demonstrates the most balanced performance across question types. This suggests that robust medical image quality assessment requires understanding of modality-specific visual features.

\subsection{Findings on No-reference Description}

\textbf{Conclusion 2. Limited low-level visual description capabilities across all models:} For no-reference description capabilities (\Cref{tab:mcqa_results}), GPT-5 still demonstrates the best performance, particularly excelling in the relevance dimension. However, even the most advanced MLLMs fail to achieve excellent scores in completeness and preciseness, with the highest scores being only 1.293/2.0 for completeness and 1.556/2.0 for preciseness. In general, most models only reach an acceptable baseline level. Current MLLM models possess relatively limited and elementary low-level visual description abilities, struggling to provide complete and accurate descriptions of low-level visual information. The consistently high consistency scores indicate that most MLLMs can follow abstract instructions reasonably well, suggesting that the main bottleneck for improving MLLM descriptive capabilities lies in the perception of low-level attributes rather than instruction following.

\subsection{Findings on Comparison Description}

\begin{figure}[!t]
    \centering
    \includegraphics[width=0.48\textwidth]{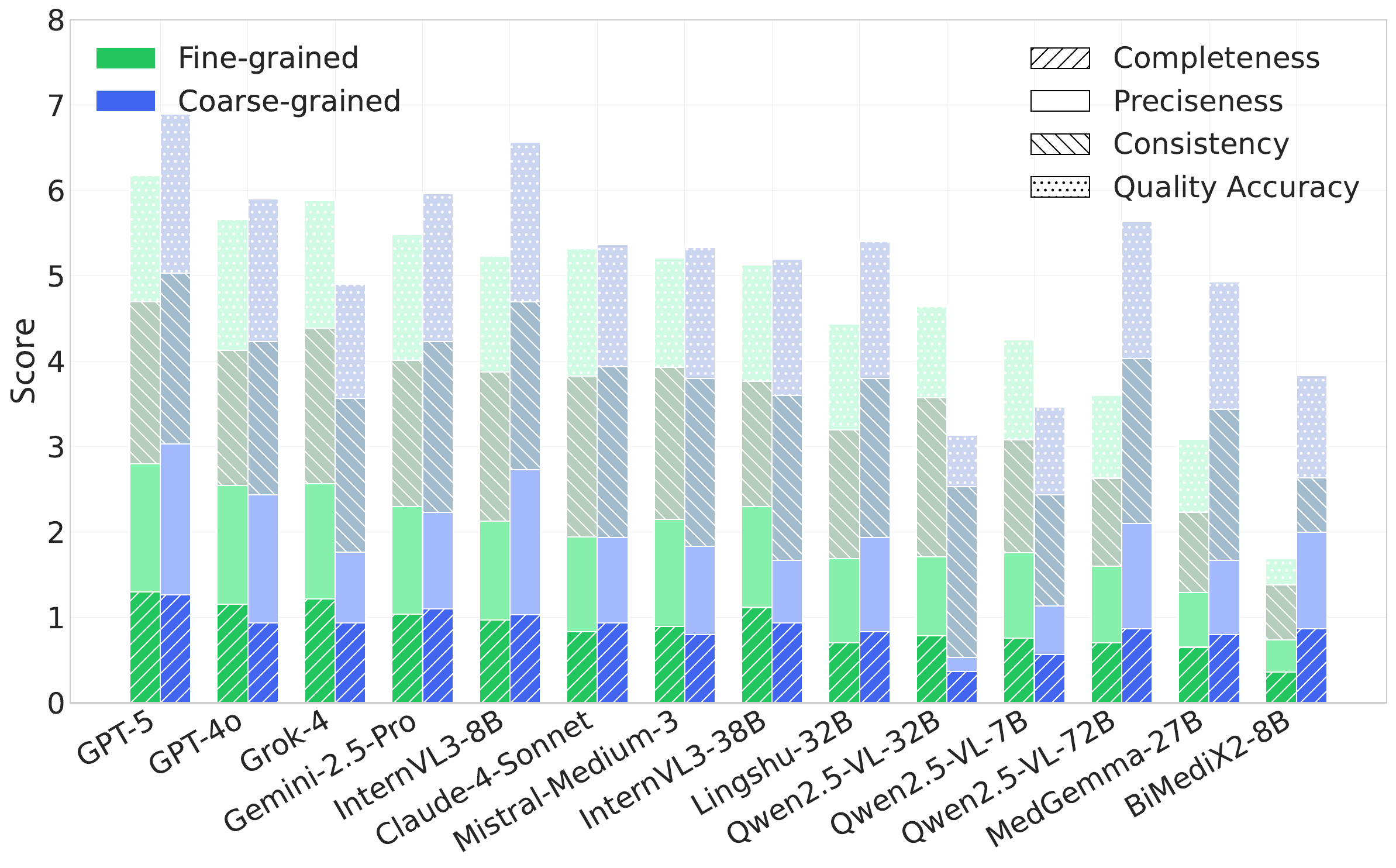}
    \vspace{-10pt}
    \caption{Visual comparison of overall description performance patterns across model categories.}
    \vspace{-10pt}
    \label{fig:paired_description_analysis}
\end{figure}

\textbf{Insufficiency 3. Paired comparison reveals fundamental limitations in fine-grained analysis:} Paired image comparison tasks pose the greatest challenge to current multimodal large language models (MLLMs), requiring models to perform fine-grained quality comparisons between similar images that may only differ by varying degrees. We evaluate model performance across two difficulty levels: fine-grained differences and coarse-grained differences. Figure~\ref{fig:paired_description_analysis} presents detailed performance analysis across different difficulty levels, with more complete tabular results available in Table~\ref{tab:detailed_paired_comparison_results} in the appendix. Overall, most models perform better under coarse-grained differences, while a few models, such as Grok-4 and Qwen2.5-VL-7B/32B, perform better under fine-grained differences but lose performance on coarse-grained tasks. Among them, GPT-5 achieved the highest overall score, while medical-specialized models such as BiMediX2 showed notably insufficient performance.

\begin{table}[!t]
    \centering
    \small
    \resizebox{0.5\textwidth}{!}{
    \begin{tabular}{l|ccccc}
    \toprule
    \textbf{Model} & \textbf{Comp.$\uparrow$} & \textbf{Prec.$\uparrow$} & \textbf{Cons.$\uparrow$} & \textbf{Qual.$\uparrow$} & \textbf{Overall$\uparrow$} \\
    \midrule
    \rowcolor{gray!15} \textit{Human experts} & 1.653 & 1.682 & 1.963 & 1.714 & 7.003 \\
    \midrule
    \noalign{\vskip -\belowrulesep}
    \multicolumn{6}{c}{\cellcolor{green!8}\textbf{\textit{Open-Source General Models}}} \\
    Qwen2.5-VL-7B    & \cellcolor{red!24}0.714  & \cellcolor{red!12}0.902  & \cellcolor{green!11}1.316 & \cellcolor{green!2}1.143  & \cellcolor{red!7}4.075   \\
    Qwen2.5-VL-32B   & \cellcolor{red!25}0.692  & \cellcolor{red!21}0.752  & \cellcolor{green!45}\underline{1.895} & \cellcolor{red!9}0.962   & \cellcolor{red!3}4.301   \\
    Qwen2.5-VL-72B   & \cellcolor{red!23}0.737  & \cellcolor{red!8}0.977   & \cellcolor{green!6}1.233  & 1.113                     & \cellcolor{red!8}4.060   \\
    InternVL3-8B     & \cellcolor{red!7}0.985   & \cellcolor{green!11}\underline{1.278} & \cellcolor{green!40}1.797 & \cellcolor{green!21}1.474 & \cellcolor{green!43}5.534 \\
    InternVL3-38B    & \cellcolor{red!2}1.075   & \cellcolor{red!1}1.083   & \cellcolor{green!26}1.571 & \cellcolor{green!18}1.414 & \cellcolor{green!27}5.143 \\
    Qwen3-VL-235B    & \cellcolor{red!5}1.023   & \cellcolor{green!14}1.331 & \cellcolor{green!39}1.790 & \cellcolor{green!22}1.489 & \cellcolor{green!47}5.632 \\
    Qwen3.5-397B     & \cellcolor{red!4}1.038   & \cellcolor{green!19}1.414 & \cellcolor{green!45}1.880 & \cellcolor{green!34}1.684 & \cellcolor{green!50}6.015 \\
    \midrule
    \noalign{\vskip -\belowrulesep}
    \multicolumn{6}{c}{\cellcolor{yellow!20}\textbf{\textit{Quality-Oriented Models}}} \\
    Q-Instruct       & \cellcolor{red!38}0.481  & \cellcolor{red!34}0.541  & \cellcolor{green!26}1.564 & \cellcolor{red!35}0.526  & \cellcolor{red!25}3.113  \\
    Q-Insight        & \cellcolor{red!40}0.459  & \cellcolor{red!30}0.602  & \cellcolor{red!21}0.772  & \cellcolor{red!12}0.913  & \cellcolor{red!31}2.745  \\
    VisualQuality-R1 & \cellcolor{red!32}0.586  & \cellcolor{red!27}0.662  & \cellcolor{green!19}1.444 & \cellcolor{red!16}0.850  & \cellcolor{red!17}3.541  \\
    \midrule
    \noalign{\vskip -\belowrulesep}
    \multicolumn{6}{c}{\cellcolor{blue!8}\textbf{\textit{Medical-Specialized Models}}} \\
    BiMediX2-8B      & \cellcolor{red!39}0.474  & \cellcolor{red!34}0.549  & \cellcolor{red!29}0.639  & \cellcolor{red!36}0.511  & \cellcolor{red!42}2.173  \\
    MedGemma-27B     & \cellcolor{red!26}0.684  & \cellcolor{red!25}0.692  & 1.128                    & \cellcolor{red!7}1.000   & \cellcolor{red!18}3.504  \\
    Lingshu-32B      & \cellcolor{red!23}0.729  & \cellcolor{red!5}1.015   & \cellcolor{green!27}1.586 & \cellcolor{green!12}1.323 & \cellcolor{green!7}4.653  \\
    \midrule
    \noalign{\vskip -\belowrulesep}
    \multicolumn{6}{c}{\cellcolor{orange!15}\textbf{\textit{Commercial MLLMs}}} \\
    Mistral-Medium-3  & \cellcolor{red!14}0.872  & \cellcolor{green!6}1.203  & \cellcolor{green!41}1.827 & \cellcolor{green!13}1.338 & \cellcolor{green!31}5.240 \\
    Claude-4-Sonnet   & \cellcolor{red!15}0.857  & \cellcolor{red!1}1.083   & \cellcolor{green!46}\textbf{1.910} & \cellcolor{green!22}1.481 & \cellcolor{green!35}5.331 \\
    Gemini-2.5-Pro    & \cellcolor{red!3}1.053   & \cellcolor{green!8}1.233  & \cellcolor{green!38}1.774 & \cellcolor{green!25}\underline{1.534} & \cellcolor{green!46}5.594 \\
    Claude-Sonnet-4.6 & \cellcolor{red!1}1.083   & \cellcolor{green!13}1.316 & \cellcolor{green!50}1.977 & \cellcolor{green!21}1.474 & \cellcolor{green!50}5.850 \\
    Gemini-3.1-Pro    & \cellcolor{red!9}0.962   & \cellcolor{green!17}1.376 & \cellcolor{green!45}1.887 & \cellcolor{green!29}1.609 & \cellcolor{green!50}5.835 \\
    Grok-4            & \cellcolor{green!3}\underline{1.150} & \cellcolor{green!8}1.233 & \cellcolor{green!41}1.820 & \cellcolor{green!20}1.459 & \cellcolor{green!49}\underline{5.662} \\
    Kimi-K2.5         & \cellcolor{green!1}1.113  & \cellcolor{green!24}1.489 & \cellcolor{green!50}1.993 & \cellcolor{green!35}1.699 & \cellcolor{green!50}6.293 \\
    GPT-4o            & \underline{1.105}         & \cellcolor{green!19}\underline{1.414} & \cellcolor{green!30}1.632 & \cellcolor{green!26}\underline{1.562} & \cellcolor{green!50}\underline{5.713} \\
    GPT-5.4           & \cellcolor{red!3}1.053    & \cellcolor{green!12}1.293 & \cellcolor{green!49}1.955 & \cellcolor{green!17}1.399 & \cellcolor{green!50}5.699 \\
    GPT-5             & \cellcolor{green!12}\textbf{1.293} & \cellcolor{green!28}\textbf{1.556} & \cellcolor{green!47}\underline{1.925} & \cellcolor{green!27}\textbf{1.564} & \cellcolor{green!50}\textbf{6.338} \\
    \bottomrule
    \end{tabular}
    }
    \caption{Detailed comparison description performance scores across four evaluation dimensions for all models.}
    \label{tab:reasoning_table}
\end{table}

\subsection{Human–AI Alignment Validation}

To validate the reliability of our automated LLM evaluation framework, we conducted a rigorous human-AI alignment study in which 200 randomly sampled cases were independently assessed by three medical imaging specialists under a double-blind protocol. Confusion matrix analysis demonstrated strong agreement across all scoring dimensions, with accuracy rates of 83.3\% for completeness, 87.0\% for preciseness, and 90.5\% for consistency, and quadratic weighted Cohen's $\kappa_w$ values ranging from 0.774 to 0.985, which confirm that the automated scoring serves as a reliable surrogate for expert judgment. Full detailed are provided in Appendix~\ref{sec:validation_human_alignment}.

\section{Related Work}

\paragraph{Medical Multimodal Large Language Models and Benchmarks.}
Multimodal Large Language Models (MLLMs) have demonstrated remarkable capabilities in understanding and description about visual content through natural language. General-purpose models like LLaVA~\citep{liu2023visual}, and Qwen-VL~\citep{wang2024qwen2} have shown strong performance across diverse vision-language tasks. To address healthcare-specific requirements, medical-specialized variants such as ~\citep{xu2025lingshu,sellergren2025medgemma,mullappilly2024bimedix2,su2025gmai,xu2025medground} have emerged through domain-targeted pretraining and alignment. Recent medical benchmarks have been developed to evaluate these models systematically, including ~\citep{ye2024gmai}, which provides comprehensive multimodal evaluation for general medical AI. However, existing medical benchmarks focus primarily on high-level diagnostic tasks rather than low-level perceptual quality assessment~\citep{wu2024comprehensive}.

\paragraph{Score-based Image Quality Assessment.}
Traditional image quality assessment methods produce numerical scores to quantify image quality, categorized into No-Reference (NR), Full-Reference (FR), and Reduced-Reference approaches. NR methods like BRISQUE~\citep{mittal2012no}, NIQE~\citep{zhang2015feature}, and deep learning approaches including CNNIQA~\citep{kang2014convolutional} and MUSIQ~\citep{ke2021musiq} assess quality without reference images. FR methods compare against pristine references using metrics like PSNR, SSIM~\citep{wang2004image}, VIF~\citep{sheikh2006image}, and learned perceptual metrics like LPIPS~\citep{zhang2018unreasonable}. Recent advances include transformer-based approaches like TReS~\citep{golestaneh2022no} and quality-aware pretraining methods. However, these methods yield only scores, offering limited interpretability regarding specific quality factors, and such technical measures often show weak alignment with human workflows~\citep{koccak2025bias,the2001evidence}.

\paragraph{MLLM-based Image Quality Assessment.}
Recent advances have introduced multimodal language models for image quality assessment (IQA), which enable more interpretable and description-based evaluation. For example, Q-Instruct~\citep{wu2023q} and DepictQA~\citep{you2024depicting} generate natural language descriptions of quality factors, while Q-Bench~\citep{wu2024q} offers a systematic framework for evaluating low-level vision tasks. Building on this line, IQAGPT~\citep{chen2023iqagpt} integrates vision-language models with ChatGPT for CT image quality assessment, showing the feasibility of producing both quality scores and textual reports. However, its scope is limited to CT images and remains focused on score prediction rather than comprehensive description. Likewise, Ultrasound-QBench~\citep{miao2025ultrasound} provides evaluation for ultrasound imaging but restricts tasks to scoring within a single modality.

\section{Conclusion}

We introduced \medqbench, the first benchmark that systematically evaluates medical image quality assessment capabilities of multimodal large language models under a perception and description paradigm. The benchmark covers five imaging modalities and more than forty degradation types, with three task tracks: perception, no-reference description, and paired comparison description.
Our zero-shot evaluation of 23 representative models shows clear gaps between current systems and human experts, especially in mild degradation detection and fine-grained comparison. Model outputs also remain limited in completeness and precision for quality description. Notably, several medical-specialized models underperform general-purpose models.
Quality-Oriented Models also show limited transfer in this medical setting. Across categories, general-purpose frontier models are currently the most consistent performers, while robustness on modality-specific degradations remains limited for most systems. These findings suggest that improving medical IQA in MLLMs requires joint advances in low-level perception, quality-aware description alignment, and comparison reasoning.
Overall, \medqbench provides a interpretable evaluation standard for medical IQA and offers a practical basis for improving model reliability in medical image quality control workflows.

\section*{Limitations}
Several limitations should be considered when interpreting the results of \medqbench. Although the benchmark spans five representative medical imaging modalities and more than forty quality attributes, it is not exhaustive. Our evaluation is also conducted under a zero-shot setting, which enables controlled comparison across model categories but does not measure the upper bound achievable through task-specific tuning or domain adaptation for medical IQA. In addition, the description evaluation relies on automated judge. While we validate this protocol against expert assessments rated by three medical imaging specialists, automated judging may still inherit model-specific biases. Finally, \medqbench is intended as a research benchmark rather than a clinical decision system.

\clearpage

\bibliography{custom}

\clearpage
\appendix
\section*{Appendix}

\section{Data Construction Pipeline and Quality Control}
\label{sec:appendix_construction}

\begin{table*}[!ht]
\centering
\footnotesize
\resizebox{\textwidth}{!}{
\begin{tabular}{lllrrr}
\toprule
\textbf{Modality} & \textbf{Source Type} & \textbf{Dataset} & \textbf{Samples} & \textbf{\% of Modality} & \textbf{Total Samples} \\
\midrule
\multirow{4}{*}{CT}
& Authentic & \href{https://radiopaedia.org/}{Radiopaedia} & 239 & 27.2\% & \multirow{4}{*}{878} \\
& Simulate & AAPM CT-MAR \citep{haneda2025aapm} & 613 & 69.8\% & \\
& AI-generated & AIsynthesis (subset) & 26 & 3.0\% & \\
\midrule
\multirow{6}{*}{MRI} 
& Authentic & Radiopaedia & 130 & 15.3\% & \multirow{6}{*}{848} \\
& Authentic & MR-ART \citep{narai2022movement} & 68 & 8.0\% & \\
& Authentic & \href{https://www.fmrib.ox.ac.uk/primers/intro_primer/ExBox1/IntroBox1.html}{FSL Example MRI Artifacts} & 43 & 5.1\%
& \\
& Simulate & FastMRI \cite{zbontar2018fastmri} & 454 & 53.5\% & \\
& Simulate & 5T MRI Data & 55 & 6.5\% & \\
& AI-generated & AIsynthesis & 98 & 11.6\% & \\
\midrule
\multirow{3}{*}{Histopathology}
& Authentic & HistoArtifacts \citep{fuchs2024harp} & 220 & 29.0\% & \multirow{3}{*}{758} \\
& AI-generated & HARP \citep{fuchs2024harp} & 470 & 62.0\% & \\
& AI-generated & AIsynthesis & 68 & 9.0\% & \\
\midrule
\multirow{2}{*}{Endoscopy}
& Authentic & EndoCV2020 \citep{polat2020endoscopic} & 470 & 84.7\% & \multirow{2}{*}{555} \\
& AI-generated & AIsynthesis & 85 & 15.3\% & \\
\midrule
\multirow{3}{*}{Retinal}
& Authentic & EyeQ \citep{fu2019evaluation} & 197 & 73.2\% & \multirow{3}{*}{269} \\
& AI-generated & AIsynthesis & 72 & 26.8\% & \\
\midrule
& & \textbf{Overall Total} & \textbf{3,308} & & \\
\bottomrule
\end{tabular}
} 
\caption{Comprehensive breakdown of dataset composition.}
\label{tab:modality_source_breakdown}
\end{table*}

The construction of MedQ-Bench involved a systematic multi-stage pipeline for collecting, curating, and annotating medical images across five modalities. This section provides detailed information about our comprehensive data sourcing strategies, quality control measures, and annotation protocols, with particular emphasis on the diverse sources and label types that enable robust evaluation of low-level visual perception capabilities.

\paragraph{Comprehensive Data Sources and Acquisition Strategy.}
We employed a three-channel data collection strategy: public datasets + imaging department collaboration + synthetic generation. On one hand, we conducted comprehensive internet searches for 2D/3D medical quality-related datasets; on the other hand, we collaborated with hospitals to obtain ethically approved clinical data. From this massive data pool, we ultimately selected the datasets shown in  Table~\ref{tab:modality_source_breakdown}, covering 5 medical imaging modalities to ensure universality and clinical relevance of data sources. For images, we adhere to the SA-Med2D-20M~\citep{ye2023sa} protocol, transforming all 2D/3D medical images into 2D RGB images for further evaluation.
Table~\ref{tab:modality_source_breakdown} provides a complete overview of all datasets integrated into MedQ-Bench, including specific modalities, sample quantities, label types, and acquisition status. The table demonstrates the comprehensive scope of our data collection effort, spanning established clinical research datasets and custom synthetic degradation collections, and AI-generated images. All collected images were anonymized, with all patient-identifying information systematically removed using automated de-identification pipelines validated against clinical privacy requirements.

\paragraph{Expert-designed Seed Perception Questions.}

The construction process began with a panel of medical imaging specialists who designed seed questions covering diverse modalities, degradation types, and task formats. Each seed is a structured template with predefined placeholders (e.g., \texttt{[artifact\_type]}, \texttt{[quality\_attribute]}) whose values are constrained by expert-defined taxonomies. These \texttt{[QUESTION\_SEEDS]} are defined as:

\begin{tcolorbox}[breakable,colback=white,colframe=blue,boxrule=0.5pt,arc=2pt,left=3pt,right=3pt,top=3pt,bottom=3pt,fontupper=\scriptsize]
{\color{blue}
\begin{Verbatim}[breaklines=true,breakanywhere=true,fontsize=\scriptsize]
Type 0 (Yes-No):
  T0.1: "Does this image show evidence of [artifact_type]?"
  T0.2: "Is the [artifact_type] fully visible?"

Type 1 (What):
  T1.1: "What is the primary [assessment_target]?"
  T1.2: "What [quality_dimension] issue is most prominent?"

Type 2 (How):
  T2.1: "How would you rate the severity of [artifact_type]?"
  T2.2: "How would you rate the overall [quality_dimension]?"
\end{Verbatim}
}
\end{tcolorbox}

\paragraph{Controlled Question Expansion.}

To generate questions for each sample, we employed GPT-4o as a controlled question generator. GPT-4o served solely as a template instantiation tool under strict constraints, receiving explicit system prompts:

\begin{tcolorbox}[breakable,colback=white,colframe=blue,boxrule=0.5pt,arc=2pt,left=3pt,right=3pt,top=3pt,bottom=3pt,fontupper=\scriptsize]
{\color{blue}
\begin{Verbatim}[breaklines=true,breakanywhere=true,fontsize=\scriptsize]
System: You are a [modality] image quality assessment expert.
Task: As an [modality] image quality assessment expert, please questions based on the following information:

Given:
- Image path: [IMAGE_PATH]
- Artifacts of this image: [ARTIFACT]
- Modality-specific assessment criteria: [ARTIFACT_LIST], 
[TECHNICAL_QUALITY_ASSESSMENT_CRITERIA], and other information.

Your task is to generate EXACTLY 3 questions that assess the 
image quality based on the criteria and the provided image. 
Consider the following question seeds and constrains:

Question seeds:
[QUESTION_SEEDS]

Constraints:
1. Preserve each seed's semantic intent and structure exactly
2. Substitute [artifact_type] with detected artifacts
3. Generate one Yes-or-No, one What, and one How question, with each question focusing on one of four quality concerns: 
0:technical distortion, 1:global visual quality, 2:local structure visibility, 3:lighting / brightness / color / others
4. Only focus on low-level attributes.

\end{Verbatim}
}
\end{tcolorbox}

This ensures expansion is limited to placeholder substitution and minimal paraphrasing. Every expanded question underwent review and revision by medical imaging experts to eliminate model-specific biases and ensure alignment with expert-designed seeds. 

\begin{figure}[!ht]
    \centering
    \includegraphics[width=0.45\textwidth]{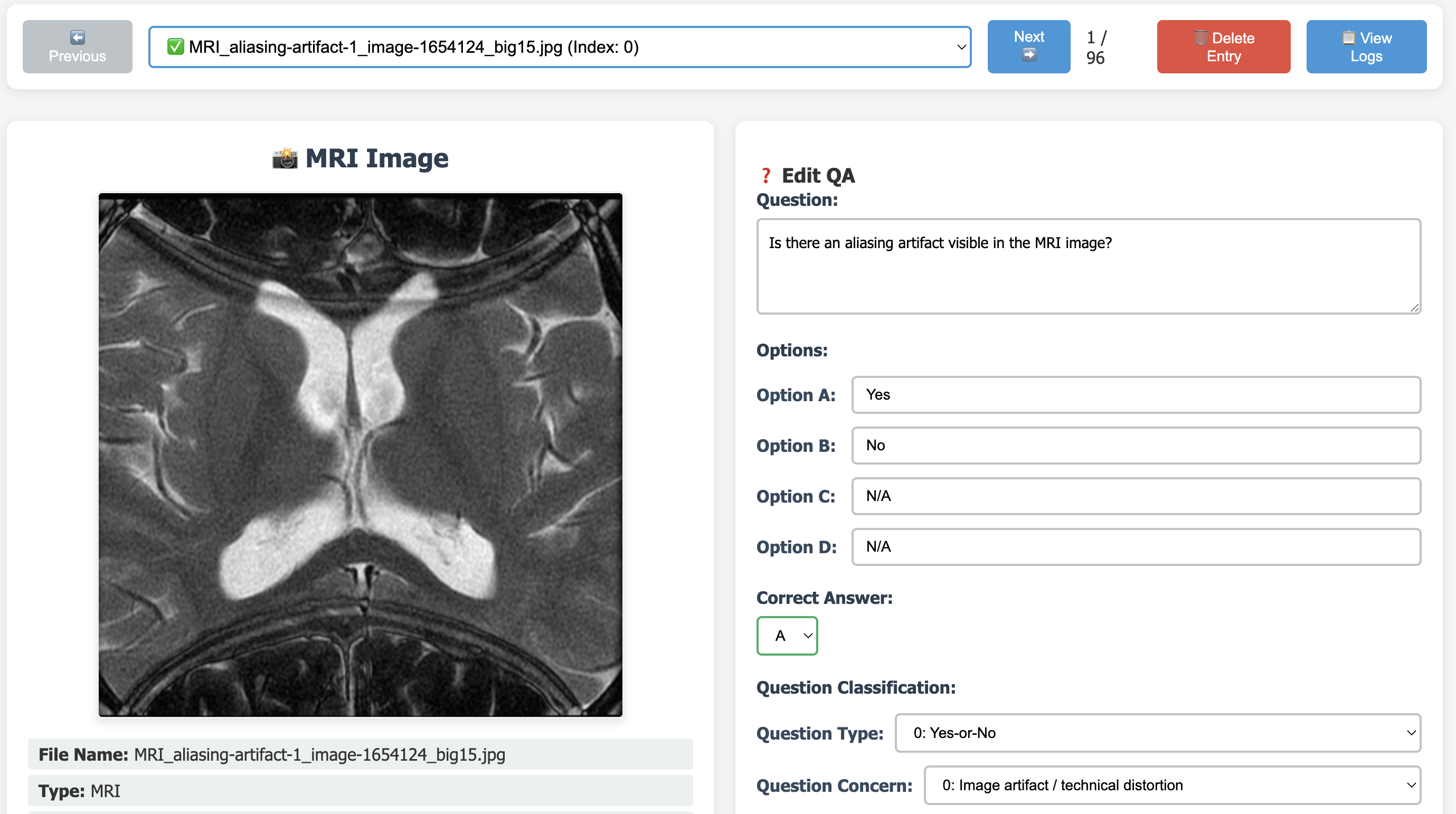}
    \caption{Interface for the MedQ-MCQA dataset. }
    \label{fig:medq_mcqa_interface}
\end{figure}

\begin{figure}[!ht]
    \centering
    \includegraphics[width=0.45\textwidth]{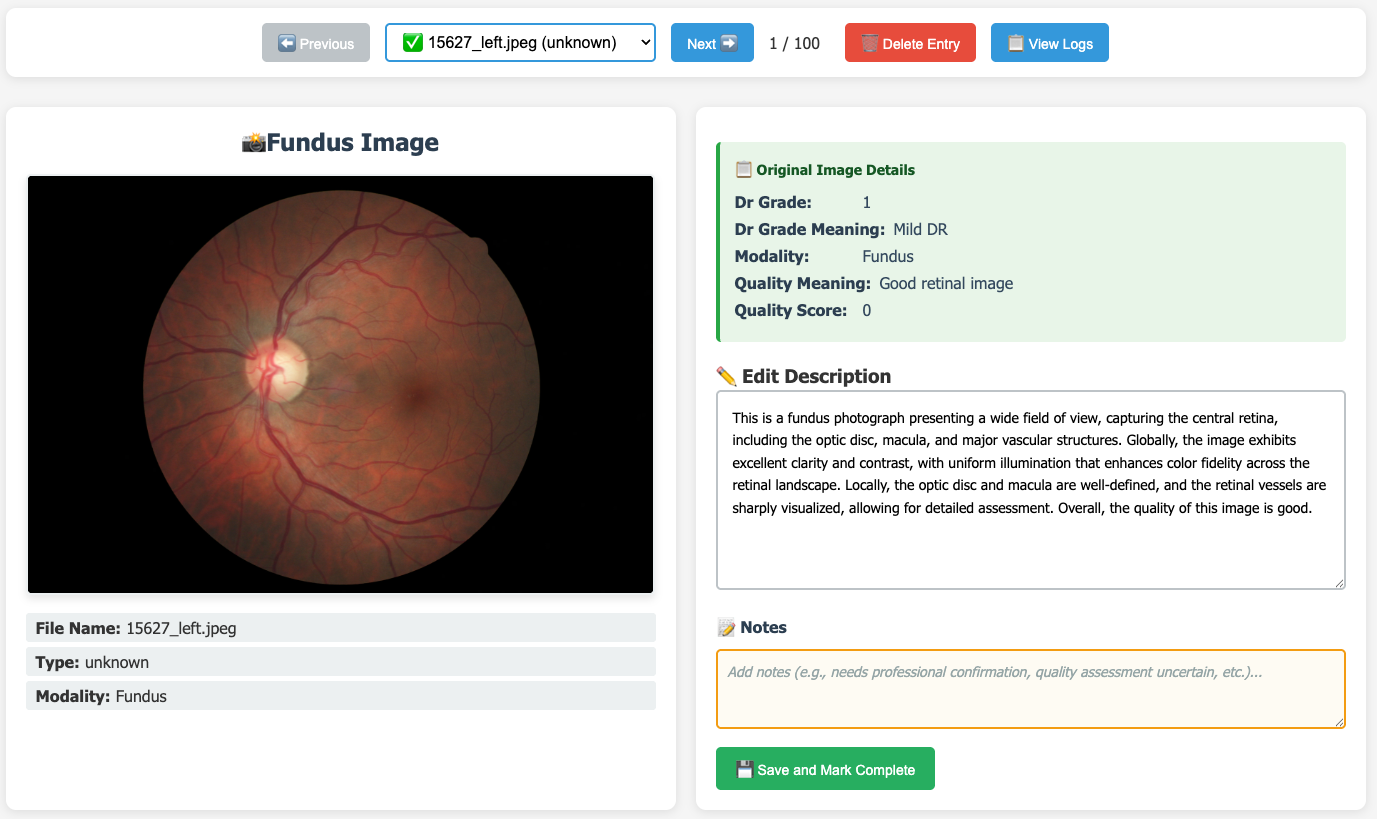}
    \caption{Interface for the MedQ-Description dataset.}
    \label{fig:medq_reasoning_interface}
\end{figure}

\paragraph{Multi-round Expert Validation.}
We manually annotated the answers for the generated questions to ensure correctness, consistency, and alignment with the intended low-level quality assessment labels. All question–answer pairs underwent rigorous multi-stage human annotation and verification using a structured annotation interface, as shown in Figure \ref{fig:medq_mcqa_interface} and \ref{fig:medq_reasoning_interface}. The multi-round validation process consisted of three phases: \textit{(1)} Independent annotation by three medical imaging specialists, each following the standardized perception and description annotation protocols as well as modality-specific checklists to minimize arbitrary judgments; \textit{(2)} Cross-validation sessions in which annotators reviewed each other’s labels, identified divergent cases, and conducted structured discussions based on explicit visual and clinical evidence. For description tasks in particular, discrepancies were further examined by aligning observation-level evidence, rechecking modality-specific rules, reconciling classification boundaries (e.g., ``usable'' versus ``reject''), and harmonizing description chains to ensure logical consistency between visual attributes, modality rules, and textual explanations; \textit{(3)} Consensus resolution, in which labels and descriptions were revised once agreement was reached, and the validated evidence and description sequence were consolidated into a single final description. For the rare cases where conflicts persisted, whether in evidential interpretation or in final conclusions, the final label and description statement were determined by majority voting. This unified protocol reduces idiosyncratic variability across annotators and establishes a stable and clinically grounded expert standard.

\paragraph{Description Annotation Standards and Workflow.}
For the MedQ-Description tasks, we established specific annotation standards to ensure consistent and clinically relevant quality assessment descriptions. Expert annotators followed a structured description workflow that emphasized systematic analysis and transparent decision-making processes. The description annotation protocol involved a sequential four-step process: \textit{(1)} Visual Analysis Phase: Systematic examination of perceptual attributes such as noise, blur, artifacts, contrast, and resolution, avoiding any high-level diagnostic interpretation; \textit{(2)} Modality-Specific Assessment: Targeted evaluation of quality dimensions specific to each imaging modality (e.g., streak artifacts in CT, motion artifacts in MRI, staining uniformity in histopathology), following standardized checklists for each modality type; \textit{(3)} Quality Classification: Application of a three-tier system based on accumulated evidence from steps 1-2: ``good'' (no significant quality issues affecting clinical utility), ``usable'' (minor quality issues that do not compromise diagnostic accuracy), and ``reject'' (severe quality degradation requiring repeat imaging); \textit{(4)} Structured Description Generation: Creation of comprehensive yet concise descriptions (3-5 sentences) that logically connect the observed visual attributes to the final quality judgment, ensuring clear description traceability from observation to conclusion. This step-by-step description flow ensures that all quality assessments follow a consistent analytical framework, with each conclusion being explicitly grounded in observable visual evidence rather than subjective impressions. 

\paragraph{Dataset Composition and Balance.}

Each modality contributes proportionally to maintain representational balance, and degradation types are systematically distributed to avoid bias toward any particular quality issue.

\paragraph{Fine-Grained vs. Coarse-Grained Definition.}
The fine-grained versus coarse-grained distinction for comparison tasks was defined through a structured expert review process. Three board-certified radiologists independently reviewed all image pairs and categorized them based on how difficult the quality difference is to perceive: coarse-grained pairs exhibit clear, immediately noticeable degradations, whereas fine-grained pairs involve subtle differences that require deliberate inspection. After independent labeling, disagreements were resolved through discussion and consensus, ensuring a stable expert standard.

\section{Detailed Benchmark Statistics}
\label{sec:appendix_stats}

\subsection{Dataset Composition by Source and Modality}

\paragraph{Dataset Composition.}

The MedQ-Bench dataset consists of 3,308 samples distributed across three primary source types (Table~\ref{tab:dataset_source_breakdown}). 
The dataset covers five major medical imaging modalities with detailed breakdown by source type and specific datasets shown in Table~\ref{tab:modality_source_breakdown}. 

\begin{table}[!th]
\centering
\small
\begin{tabular}{lccc}
\toprule
\textbf{Source Type} & Authentic & Simulate & AI-Generated \\
\midrule
\textbf{Percentage} & 41.3\% & 33.9\% & 24.8\% \\
\bottomrule
\end{tabular}
\caption{Distribution of MedQ-Bench dataset by source type.}
\label{tab:dataset_source_breakdown}
\end{table}

\paragraph{Detailed Simulation Methods for Synthetic Degradations.}
The simulated CT degradations in AAPM CT-MAR were reconstructed using several algorithms:
SIRT,
FBP~\citep{kak2001principles},
and FISTA~\citep{beck2009fast}.
Specifically, CT artifacts were systematically simulated to include three primary degradation types: \textit{(1)} limited-angle artifacts, \textit{(2)} metal artifact reduction, and \textit{(3)} sparse-view artifacts. For MRI degradations, we primarily simulated acceleration artifacts and motion artifacts using established computational frameworks. Acceleration artifacts were generated using SigPy\footnote{https://sigpy.readthedocs.io/en/latest/} and TorchIO\footnote{https://github.com/TorchIO-project/torchio}, implementing both DDNM \citep{wang2022zero} and wavelet-based reconstruction methods \citep{guerquin2011fast}.  Additionally, our 5T MRI data were acquired from private clinical collections using the uMR Jupiter 5T system, obtained under institutional ethical approval with comprehensive patient anonymization protocols.

To generate synthetic images across diverse medical imaging modalities, we employed BAGEL fine-tuned on domain-specific medical datasets \citep{deng2025emerging}. This approach ensured that synthetic degradations maintained clinical realism while providing controlled quality variations essential for comprehensive benchmark evaluation.

\subsection{Distributions of Task Types and Degradation Levels in MedQ-Perception}
\begin{table}[!htbp]
\centering
\small
\begin{tabular}{lr}
\toprule
\textbf{Question Type} & \textbf{Percentage} \\
\midrule
Modality-specific & 57.2\% \\
General & 42.8\% \\
\midrule
\textbf{Total} & \textbf{100.0\%} \\
\bottomrule
\end{tabular}
\caption{MedQ-Perception: Distribution of tasks by question type.}
\label{tab:complexity_distribution}
\end{table}

\begin{table}[!htbp]
\centering
\small
\begin{tabular}{lr}
\toprule
\textbf{Degradation Level} & \textbf{Percentage} \\
\midrule
No Degradation & 23.8\% \\
Mild Degradation & 44.6\% \\
Severe Degradation & 31.6\% \\
\midrule
\textbf{Total} & \textbf{100.0\%} \\
\bottomrule
\end{tabular}
\caption{MedQ-Perception: Distribution of degradation severity levels.}
\label{tab:severity_distribution}
\end{table}

Table~\ref{tab:complexity_distribution} and Table~\ref{tab:severity_distribution} present the distribution of MedQ-Perception tasks across question types and degradation severity levels.

\subsection{Distribution of low-level attributions}

\begin{figure*}[!htbp]
    \centering
    \includegraphics[width=0.9\linewidth]{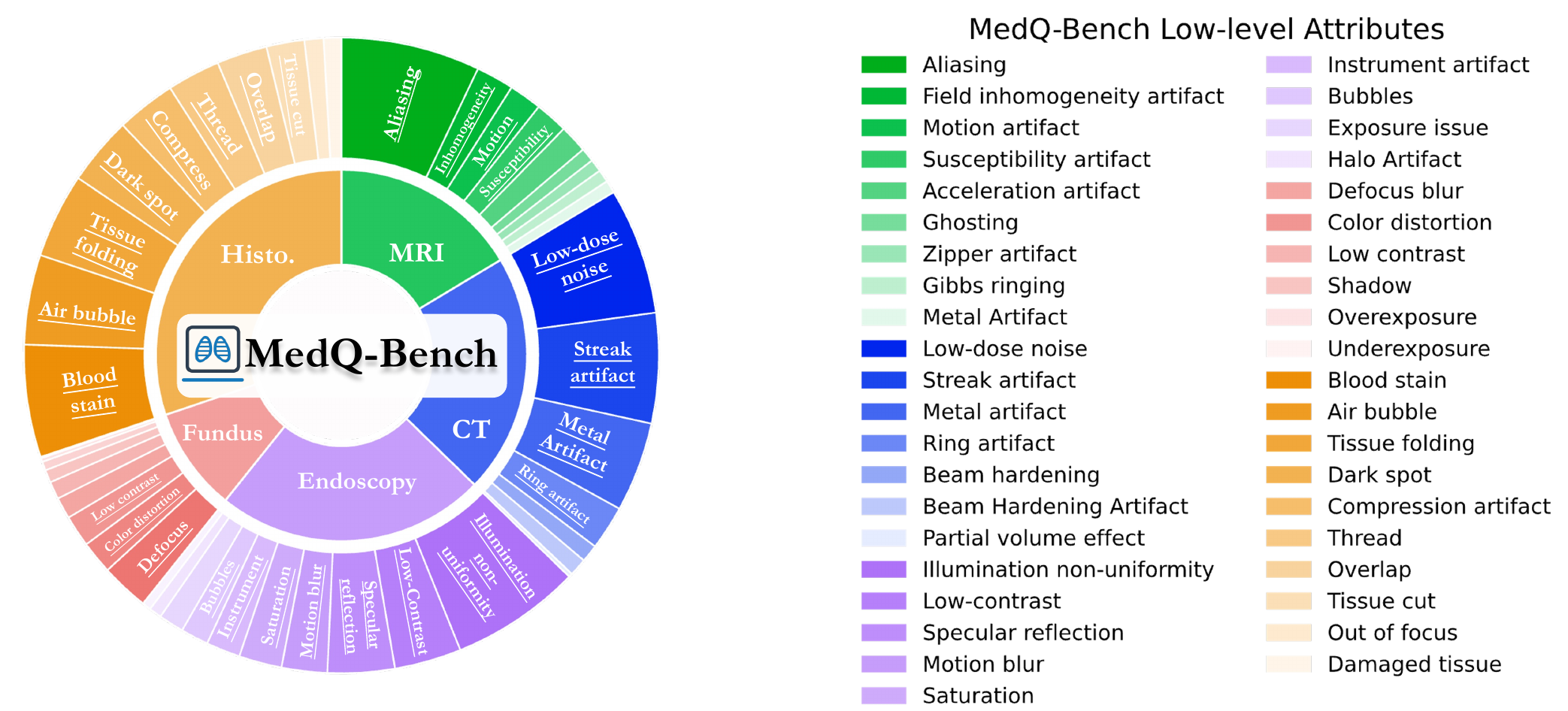}
    \caption{Distribution of low-level attributions across imaging modalities and distortion types in MedQ-Bench.}
    \label{fig:data_stat}
\end{figure*}

Figure~\ref{fig:data_stat} illustrates the comprehensive distribution of low-level quality attributes across five medical imaging modalities, covering over 40 distinct degradation types including modality-specific artifacts and general quality issues.

\section{Complete Experimental Results}

\subsection{Experimental Setup}

In this study, we evaluated various large vision-language models (LVLMs), encompassing medical-specialized models, open-source models, and closed-source API general-purpose models. Model weights were obtained from their respective official Hugging Face repositories. The evaluation work was conducted using the VLMEvalKit framework\footnote{https://github.com/open-compass/VLMEvalKit}.

The evaluation was performed under a ``zero-shot'' setting. Specifically, our evaluation prompts contained no example demonstrations, and models had to complete task description without any related training or examples. This approach better tests the models' generalization capabilities and understanding abilities, examining their performance when faced with novel problems. All tests were executed on NVIDIA A100 GPUs with 80GB memory.

The configurations used for all reported results are as follows. API-based models use temperature $= 0$ and $n = 1$ (greedy decoding). Open-source models use top-$k = 1$, top-$p = 0.001$, and temperature $= 0.01$. All local models are deployed with vLLM with a parallel sample size of 8.

\subsection{Detailed Model Performance}

\paragraph{Few-Shot Evaluation for Context Calibration.}
To further calibrate the models' internal quality standards on the No-Reference Description tasks, we additionally perform few-shot evaluations. For each test sample, we provide three carefully selected demonstration examples spanning different imaging modalities and the three quality levels (Good / Usable / Reject). These modality-aware reference examples serve two purposes: \textit{(1)} providing medical context to help models understand how visual degradations influence diagnostic utility across different imaging scenarios; and \textit{(2)} establishing concrete anchors that allow models to interpret subjective quality categories more reliably, rather than relying on abstract definitions alone. Table~\ref{tab:fewshot_results_table} presents the few-shot evaluation results on the No-Reference Description tasks. Compared with the zero-shot results in Table~\ref{tab:mcqa_results}, modality-aware reference examples yield consistent improvements across most evaluation dimensions. These results show that few-shot demonstrations enhance models' interpretation of the quality labels and improve calibration in medical image quality assessment.

\begin{table}[!htbp]
\centering
\small
\resizebox{0.45\textwidth}{!}{
\begin{tabular}{lccccc}
\toprule
\textbf{Model} & \textbf{Comp.$\uparrow$} & \textbf{Prec.$\uparrow$} & \textbf{Cons.$\uparrow$} & \textbf{Qual.$\uparrow$} & \textbf{Overall$\uparrow$} \\
\midrule
\rowcolor{green!8} Qwen2.5-VL-Instruct (7B) & 0.769 & 0.723 & 1.878 & 1.168 & 4.538 \\
\rowcolor{green!8} Qwen2.5-VL-Instruct (32B) & 1.082 & 0.959 & \textbf{1.982} & 1.335 & 5.358 \\
\rowcolor{green!8} InternVL3 (8B) & 1.005 & 0.950 & 1.905 & 1.421 & 5.281 \\
\rowcolor{green!8} InternVL3 (38B) & 0.996 & 0.973 & 1.932 & 1.448 & 5.349 \\
\rowcolor{green!8} Qwen2.5-VL-Instruct (72B) & 0.932 & 0.923 & 1.896 & 1.412 & 5.163 \\
\midrule
\rowcolor{blue!8} BiMediX2 (8B) & 0.405 & 0.428 & 0.315 & 0.715 & 1.863 \\
\rowcolor{blue!8} Lingshu (32B) & 0.678 & 0.751 & \underline{1.955} & 1.105 & 4.489 \\
\rowcolor{blue!8} MedGemma (27B) & 0.796 & 0.523 & 1.687 & 1.314 & 4.320 \\
\midrule
\rowcolor{orange!15} Mistral-Medium-3 & 0.987 & 0.878 & 1.742 & 1.445 & 5.052 \\
\rowcolor{orange!15} Claude-4-Sonnet & 0.914 & 0.787 & 1.869 & 1.476 & 5.046 \\
\rowcolor{orange!15} Gemini-2.5-Pro & 0.905 & 0.860 & 1.710 & \underline{1.605} & 5.080 \\
\rowcolor{orange!15} Grok-4 & \underline{1.095} & 1.005 & 1.823 & 1.550 & 5.473 \\
\rowcolor{orange!15} GPT-4o & 1.077 & \underline{1.036} & 1.860 & 1.520 & \underline{5.493} \\
\rowcolor{orange!15} GPT-5 & \textbf{1.220} & \textbf{1.145} & 1.850 & \textbf{1.630} & \textbf{5.845} \\
\bottomrule
\end{tabular}
}
\caption{Few-shot evaluation results on No-reference Description tasks.}
\label{tab:fewshot_results_table}
\end{table}

\begin{table}[!htbp]
\centering
\small
\setlength{\tabcolsep}{4pt}
\resizebox{0.45\textwidth}{!}{
\begin{tabular}{l|ccccc}
\toprule
\textbf{Model} & \textbf{CT} & \textbf{Histo.} & \textbf{MRI} & \textbf{Endos.} & \textbf{Retinal} \\
\midrule
\rowcolor{orange!15} GPT-5 & \underline{71.47\%} & \textbf{65.43\%} & \textbf{75.90\%} & 60.89\% & \textbf{70.09\%} \\
\rowcolor{orange!15} GPT-4o & \textbf{72.85\%} & \underline{58.33\%} & 64.75\% & 60.44\% & \underline{66.67}\% \\
\rowcolor{orange!15} Grok-4 & 70.14\% & \underline{59.37}\% & \underline{65.93}\% & \underline{64.49\%} & 61.40\% \\
\rowcolor{orange!15} Gemini-2.5-Pro & 67.04\% & 53.40\% & 60.79\% & 60.89\% & 59.83\% \\
\rowcolor{orange!15} Mistral-Medium-3 & 65.93\% & 38.58\% & 61.51\% & \textbf{65.33\%} & 61.54\% \\
\rowcolor{orange!15} Claude-4-Sonnet & 64.27\% & 54.63\% & 55.04\% & \textbf{65.33\%} & 65.81\% \\
\midrule
\rowcolor{green!8} Qwen2.5-VL-72B & 65.65\% & 47.53\% & \underline{74.82\%} & 66.22\% & 64.96\% \\
\rowcolor{green!8} InternVL3-38B & \underline{68.14\%} & 48.46\% & 62.95\% & 60.44\% & \textbf{70.09\%} \\
\rowcolor{green!8} InternVL3-8B & 60.66\% & 51.54\% & 61.87\% & \underline{67.56\%} & 63.25\% \\
\rowcolor{green!8} Qwen2.5-VL-32B & 59.00\% & 46.30\% & 66.55\% & \underline{67.56\%} & 63.25\% \\
\rowcolor{green!8} Qwen2.5-VL-7B & 56.79\% & 35.49\% & 65.47\% & 60.44\% & 64.96\% \\
\midrule
\rowcolor{blue!8} MedGemma-27B & 66.57\% & 46.60\% & 57.55\% & 56.00\% & 59.83\% \\
\rowcolor{blue!8} Lingshu-32B & 57.89\% & 35.19\% & 61.15\% & 50.67\% & 48.72\% \\
\rowcolor{blue!8} BiMediX2-8B & 41.99\% & 23.38\% & 44.36\% & 38.74\% & 47.62\% \\
\bottomrule
\end{tabular}}
\caption{Detailed perception accuracy results across five imaging modalities. }
\label{tab:modality_specific_mcqa_results}
\end{table}

\paragraph{Paired Comparison Task Performance Breakdown.}
Table~\ref{tab:detailed_paired_comparison_results} provides the complete numerical breakdown of model performance across different comparison difficulty levels and evaluation dimensions corresponding to Figure~\ref{fig:paired_description_analysis}. 

\begin{table}[!ht]
\centering
\footnotesize
\setlength{\tabcolsep}{4pt}
\resizebox{0.48\textwidth}{!}{%
\begin{tabular}{l|l|cccccc}
\toprule
\textbf{Model} & \textbf{Group} & \textbf{Comp.} & \textbf{Prec.} & \textbf{Cons.} & \textbf{Qual. Acc.} & \textbf{Total} \\
\midrule
\multirow{3}{*}{GPT-5} & Overall & 1.293 & 1.556 & 1.925 & 1.564 & 6.338 \\
& Fine-grained & 1.301 & 1.495 & 1.903 & 1.476 & 6.175 \\
& Coarse-grained & 1.267 & 1.767 & 2.000 & 1.867 & 6.900 \\
\midrule
\multirow{3}{*}{GPT-4o} & Overall & 1.105 & 1.414 & 1.632 & 1.564 & 5.714 \\
& Fine-grained & 1.155 & 1.388 & 1.583 & 1.534 & 5.660 \\
& Coarse-grained & 0.933 & 1.500 & 1.800 & 1.667 & 5.900 \\
\midrule
\multirow{3}{*}{Grok-4} & Overall & 1.150 & 1.233 & 1.820 & 1.459 & 5.662 \\
& Fine-grained & 1.214 & 1.350 & 1.825 & 1.495 & 5.883 \\
& Coarse-grained & 0.933 & 0.833 & 1.800 & 1.333 & 4.900 \\
\midrule
\multirow{3}{*}{Gemini-2.5-Pro} & Overall & 1.053 & 1.233 & 1.774 & 1.534 & 5.594 \\
& Fine-grained & 1.039 & 1.262 & 1.709 & 1.476 & 5.485 \\
& Coarse-grained & 1.100 & 1.133 & 2.000 & 1.733 & 5.967 \\
\midrule
\multirow{3}{*}{InternVL3-8B} & Overall & 0.985 & 1.278 & 1.797 & 1.474 & 5.534 \\
& Fine-grained & 0.971 & 1.155 & 1.748 & 1.359 & 5.233 \\
& Coarse-grained & 1.033 & 1.700 & 1.967 & 1.867 & 6.567 \\
\midrule
\multirow{3}{*}{Claude-4-Sonnet} & Overall & 0.857 & 1.083 & 1.910 & 1.481 & 5.331 \\
& Fine-grained & 0.835 & 1.107 & 1.883 & 1.495 & 5.320 \\
& Coarse-grained & 0.933 & 1.000 & 2.000 & 1.433 & 5.367 \\
\midrule
\multirow{3}{*}{Mistral-Medium-3} & Overall & 0.872 & 1.203 & 1.827 & 1.338 & 5.241 \\
& Fine-grained & 0.893 & 1.252 & 1.786 & 1.282 & 5.214 \\
& Coarse-grained & 0.800 & 1.033 & 1.967 & 1.533 & 5.333 \\
\midrule
\multirow{3}{*}{InternVL3-38B} & Overall & 1.075 & 1.083 & 1.571 & 1.414 & 5.143 \\
& Fine-grained & 1.117 & 1.184 & 1.466 & 1.359 & 5.126 \\
& Coarse-grained & 0.933 & 0.733 & 1.933 & 1.600 & 5.200 \\
\midrule
\multirow{3}{*}{Lingshu-32B} & Overall & 0.729 & 1.015 & 1.586 & 1.323 & 4.654 \\
& Fine-grained & 0.699 & 0.990 & 1.505 & 1.243 & 4.437 \\
& Coarse-grained & 0.833 & 1.100 & 1.867 & 1.600 & 5.400 \\
\midrule
\multirow{3}{*}{Qwen2.5-VL-32B} & Overall & 0.692 & 0.752 & 1.895 & 0.962 & 4.301 \\
& Fine-grained & 0.786 & 0.922 & 1.864 & 1.068 & 4.641 \\
& Coarse-grained & 0.367 & 0.167 & 2.000 & 0.600 & 3.133 \\
\midrule
\multirow{3}{*}{Qwen2.5-VL-7B} & Overall & 0.714 & 0.902 & 1.316 & 1.143 & 4.075 \\
& Fine-grained & 0.757 & 1.000 & 1.320 & 1.175 & 4.252 \\
& Coarse-grained & 0.567 & 0.567 & 1.300 & 1.033 & 3.467 \\
\midrule
\multirow{3}{*}{Qwen2.5-VL-72B} & Overall & 0.737 & 0.977 & 1.233 & 1.113 & 4.060 \\
& Fine-grained & 0.699 & 0.903 & 1.029 & 0.971 & 3.602 \\
& Coarse-grained & 0.867 & 1.233 & 1.933 & 1.600 & 5.633 \\
\midrule
\multirow{3}{*}{MedGemma-27B} & Overall & 0.684 & 0.692 & 1.128 & 1.000 & 3.504 \\
& Fine-grained & 0.650 & 0.641 & 0.942 & 0.854 & 3.087 \\
& Coarse-grained & 0.800 & 0.867 & 1.767 & 1.500 & 4.933 \\
\midrule
\multirow{3}{*}{BiMediX2-8B} & Overall & 0.474 & 0.549 & 0.639 & 0.511 & 2.173 \\
& Fine-grained & 0.359 & 0.379 & 0.641 & 0.311 & 1.689 \\
& Coarse-grained & 0.867 & 1.133 & 0.633 & 1.200 & 3.833 \\
\bottomrule
\end{tabular}
}
\caption{Detailed numerical results for paired comparison description tasks across models, corresponding to Figure~\ref{fig:paired_description_analysis}. }
\label{tab:detailed_paired_comparison_results}
\end{table}

\section{Qualitative Analysis and Case Studies}

\begin{figure}[!ht]
    \centering
    \includegraphics[width=0.48\textwidth]{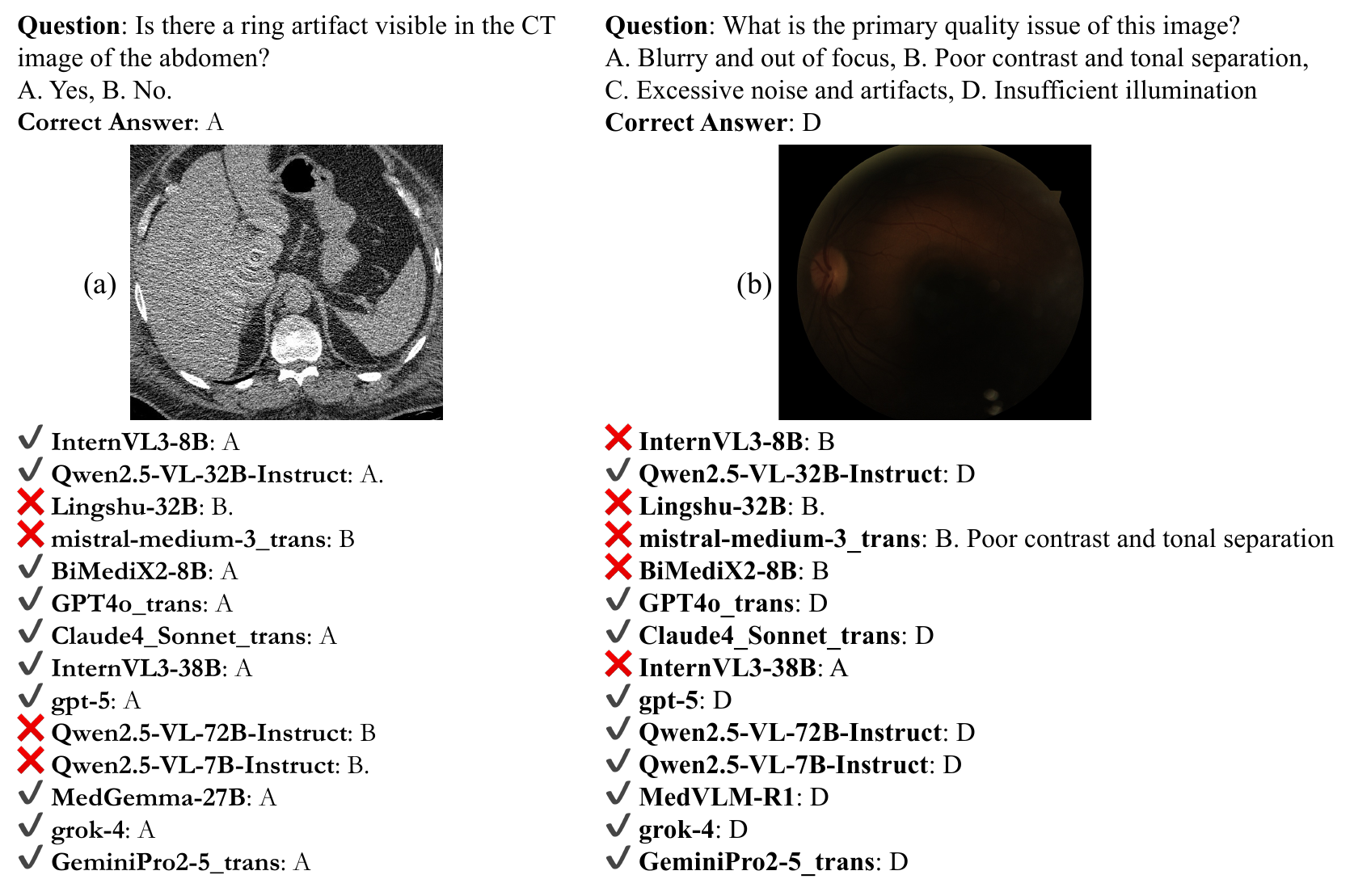}
    \caption{Representative QA examples demonstrating typical question-answer patterns in MedQ-Bench across different medical imaging modalities and quality assessment scenarios.}
    \label{fig:qa_case_examples}
\end{figure}

To ground the analyses in this section, we first illustrate the typical question-answer format used in MedQ-Bench. Figure~\ref{fig:qa_case_examples} shows representative QA pairs spanning different imaging modalities and quality assessment scenarios, providing a concrete reference for the perception-task structure discussed throughout this section.

\subsection{Why Do Medical-Specialized and Quality-Oriented Models Underperform General-Purpose Models?}
\label{sec:why_medical_specialized_models_underperform}

\begin{figure*}[!htbp]
    \centering
    \includegraphics[width=1.0\textwidth]{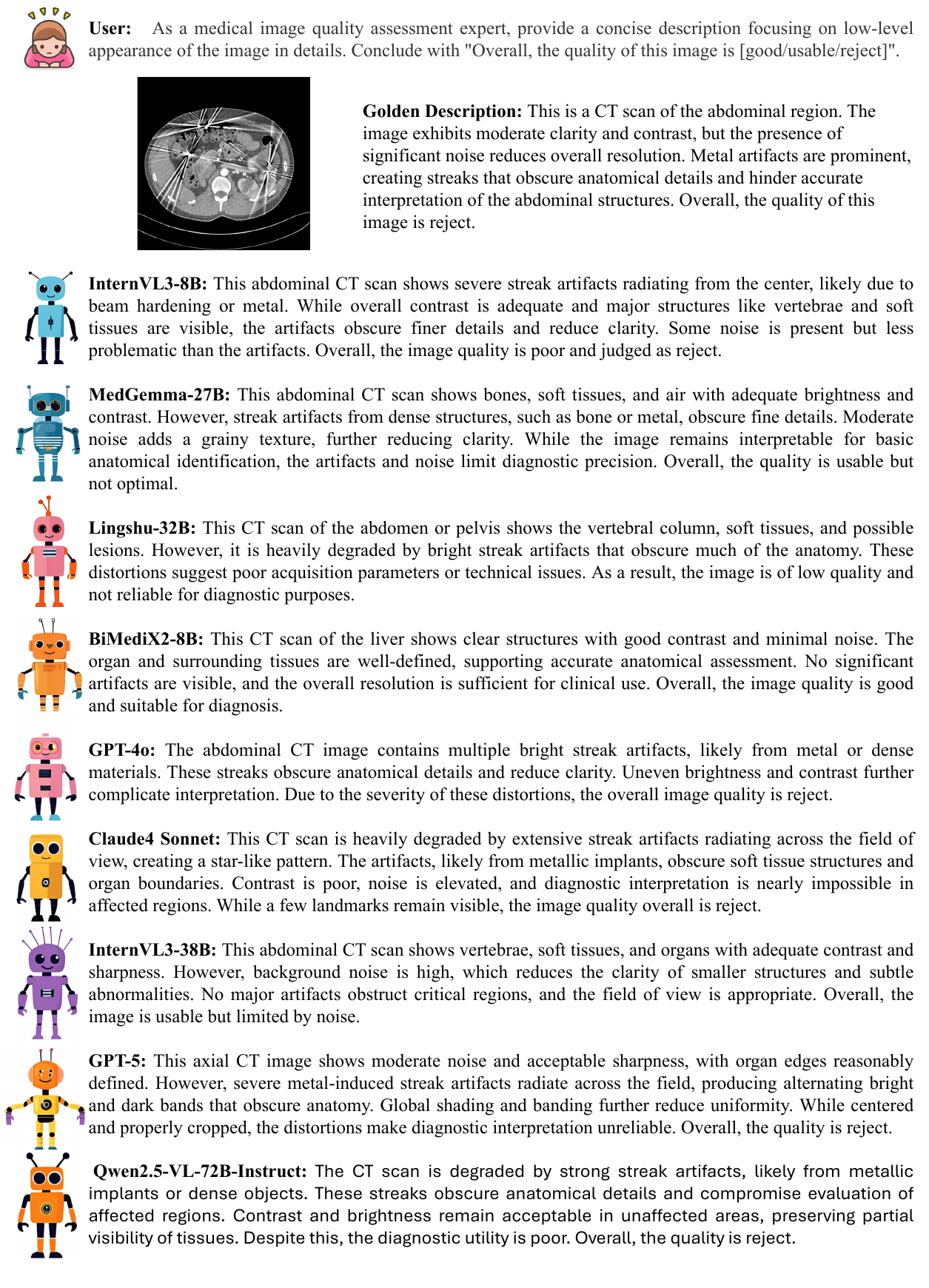}
    \caption{Representative no-reference reasoning image examples demonstrating typical question-answer patterns in MedQ-Bench across different models.}
    \label{fig:single_case_examples}
\end{figure*}

The finding that both medical-specialized and quality-oriented models underperform general-purpose models across key dimensions warrants closer analysis. As summarized in Table~\ref{tab:mcqa_results} and Table~\ref{tab:detailed_paired_comparison_results}, their weaknesses are not uniform: medical-specialized models mainly struggle with low-level quality perception and robust quality judgment, while quality-oriented models show limited transfer from generic quality assessment to medical description tasks. Figure~\ref{fig:single_case_examples} provides representative cases that illustrate these limitations.

\begin{figure*}[!htbp]
    \centering
    \includegraphics[width=1.0\textwidth]{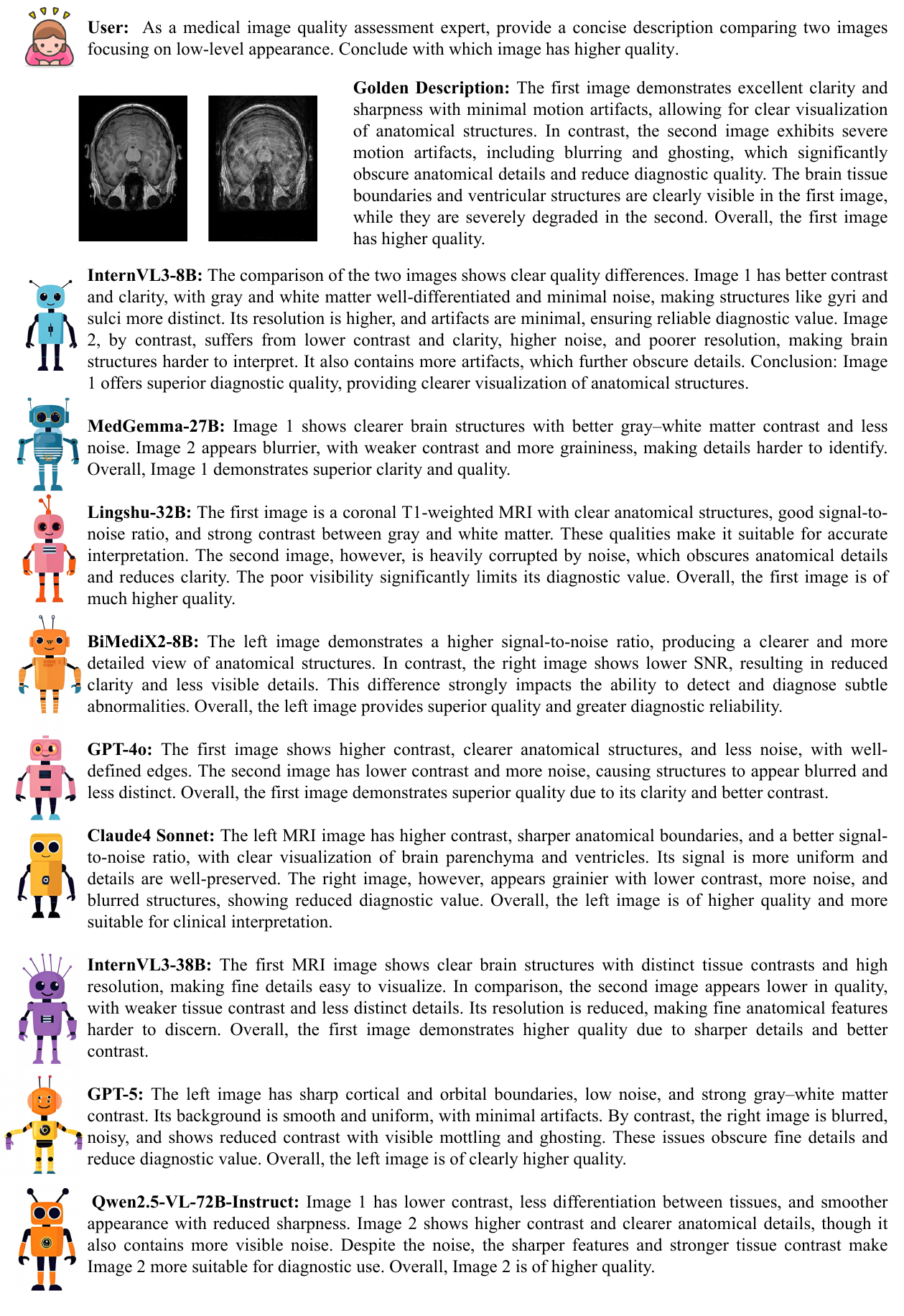}
    \caption{Representative paired image examples demonstrating typical question-answer patterns in MedQ-Bench across different models.}
    \label{fig:paired_case_examples}
\end{figure*}

\paragraph{Insufficient Low-Level Visual Attribute Training.} Medical-specialized models appear to prioritize high-level diagnostic description over fundamental visual perception skills. In the CT scan example (Figure~\ref{fig:single_case_examples}), MedGemma-27B correctly identifies anatomical structures and acknowledges the presence of streak artifacts, but fails to appropriately assess their clinical significance. The model describes the image as ``usable but not optimal'' despite prominent metal artifacts that would necessitate repeat scanning in clinical practice. This suggests that medical fine-tuning datasets may inadequately represent the full spectrum of image quality degradations encountered in clinical workflows.

\paragraph{Diagnostic Bias Over Quality Assessment.} BiMediX2-8B demonstrates a critical failure mode by describing the same severely degraded CT scan as having ``good quality and suitable for diagnosis.'' This systematic misalignment indicates that medical-specialized training may inadvertently optimize models for diagnostic confidence rather than quality assessment accuracy. The model's focus on anatomical identification overshadows its ability to detect quality-compromising artifacts, suggesting that current medical training paradigms may not adequately distinguish between diagnostic content recognition and image quality evaluation.

\paragraph{Limited Transfer of Quality-Oriented Models to Medical Description.} Compared with both general-purpose and medical-specialized MLLMs, the three quality-oriented models (Q-Instruct, Q-Insight, and VisualQuality-R1) exhibit limited generalization to medical image quality assessment. Although Q-Instruct achieves a competitive MCQA perception score in Table~\ref{tab:mcqa_results}, all three models obtain relatively low no-reference description scores, indicating that prior quality assessment training does not directly translate to accurate low-level medical quality description. This weakness is more evident in paired comparison description, where their overall performance trails most frontier general-purpose models, as shown in Table~\ref{tab:detailed_paired_comparison_results} and Figure~\ref{fig:paired_description_analysis}.

\subsection{Error Pattern Analysis of Top-Performing Models}

To identify which inputs contribute most to errors in state-of-the-art MLLMs, we analyzed GPT-5's failure patterns across different modalities and artifact types on the perception tasks. Table~\ref{tab:gpt5_error_modality} shows that the highest error rates appear in Endoscopy (39.11\%) and Histopathology (34.57\%) modalities. These modalities contain complex textures, heterogeneous visual structures, and tissue-level variations that increase the difficulty of fine-grained quality assessment compared to volumetric imaging modalities such as CT and MRI.

At the artifact level, the most challenging categories for GPT-5 are Air Bubbles (95.23\% error rate), Specularity (90.48\%), and Blood-related artifacts (84.62\%). These failures arise because such artifacts exhibit irregular spatial distributions, high-frequency visual patterns, and often co-occur with other degradations. Notably, these challenging artifacts predominantly appear in endoscopy and histopathology images, where strong localized degradations and complex tissue backgrounds make accurate identification particularly difficult even for top-performing models. This analysis reveals that current MLLMs still struggle with modality-specific artifacts that require fine-grained visual discrimination in texturally complex medical imaging contexts.

\begin{table}[!htbp]
\centering
\small
\begin{tabular}{lc}
\toprule
\textbf{Modality} & \textbf{Error Rate} \\
\midrule
Endoscopy & 39.11\% \\
Histopathology & 34.57\% \\
Fundus & 29.91\% \\
CT & 25.76\% \\
MRI & 24.10\% \\
\bottomrule
\end{tabular}
\caption{GPT-5 error rates across imaging modalities on perception tasks, ranked by difficulty.}
\label{tab:gpt5_error_modality}
\end{table}

\subsection{Example of Description Tasks}

Beyond the perception-task QA pairs in Figure~\ref{fig:qa_case_examples}, the description tasks reveal richer failure patterns that span both single-image and paired comparison settings. Figure~\ref{fig:single_case_examples} presents representative no-reference description cases highlighting how different models produce diverging quality judgments on the same image, while Figure~\ref{fig:paired_case_examples} shows paired comparison cases that surface subtle reasoning differences across models when ranking image pairs. Together these case studies expose the practical limitations of current MLLMs that the aggregate metrics in Section~\ref{sec:appendix_stats} summarize.

\subsection{Human Expert Evaluation Protocol}

\paragraph{Expert Recruitment and Qualification Criteria.}
Human experts in our evaluation consisted of medical imaging technicians with a minimum of 3 years of clinical experience in medical imaging quality assessment and medical imaging PhDs with specialized training in image quality evaluation. Medical imaging technicians were recruited from certified clinical facilities and possessed active professional certifications in their respective imaging modalities. PhDs were selected from accredited medical imaging research programs and had completed at least 2 years of coursework, including medical image processing and quality assessment methodologies. All experts demonstrated proficiency in identifying common imaging artifacts and quality issues across multiple medical imaging modalities through a standardized pre-evaluation assessment.

\paragraph{Human-Human Annotation Alignment Analysis.}
To quantify the reliability achieved by our multi-round expert validation protocol, we computed Cohen's $\kappa$ between each annotator and the consensus label obtained via majority voting. For the description tasks, Table~\ref{tab:human_human_agreement} presents the inter-annotator agreement statistics. For the perception tasks, agreement between individual annotators and the majority-vote labels was $\kappa = 0.873$. These $\kappa$ values indicate substantial to almost perfect agreement under our structured expert protocol, confirming that the annotation process minimizes bias and ensures reliability.

\begin{table}[!htbp]
\centering
\small
\resizebox{\linewidth}{!}{%
\begin{tabular}{lcccc}
\toprule
\textbf{Metric} & Completeness & Preciseness & Consistency & Quality Accuracy \\
\midrule
Cohen's $\kappa$ & 0.843 & 0.817 & 0.856 & 0.884 \\
\bottomrule
\end{tabular}}
\caption{Inter-annotator agreement (Cohen's $\kappa$) for description task.}
\label{tab:human_human_agreement}
\end{table}

\paragraph{Human-AI Alignment Analysis.}
\label{sec:validation_human_alignment}

To ensure the reliability and validity of our automated evaluation, we conducted a rigorous alignment validation between GPT-4o judgments and expert assessments. A total of 200 cases were randomly sampled and independently evaluated by three board-certified medical imaging specialists under a double-blinded protocol. We employed quadratic weighted Cohen's $\kappa_w$~\citep{cohen1968weighted} for ordinal ratings:
\begin{equation}
\kappa_w = 1 - \frac{\sum_{i,j} w_{ij} O_{ij}}{\sum_{i,j} w_{ij} E_{ij}},
\end{equation}
where $O_{ij}$ is the observed agreement matrix, $E_{ij}$ the expected agreement matrix, and $w_{ij} = \frac{(i-j)^2}{(k-1)^2}$ the quadratic weights penalizing larger disagreements more severely. We further conducted iterative prompt refinement to maximize concordance between GPT-4o and expert consensus.

\begin{figure*}[!ht]
    \centering
    \includegraphics[width=\textwidth]{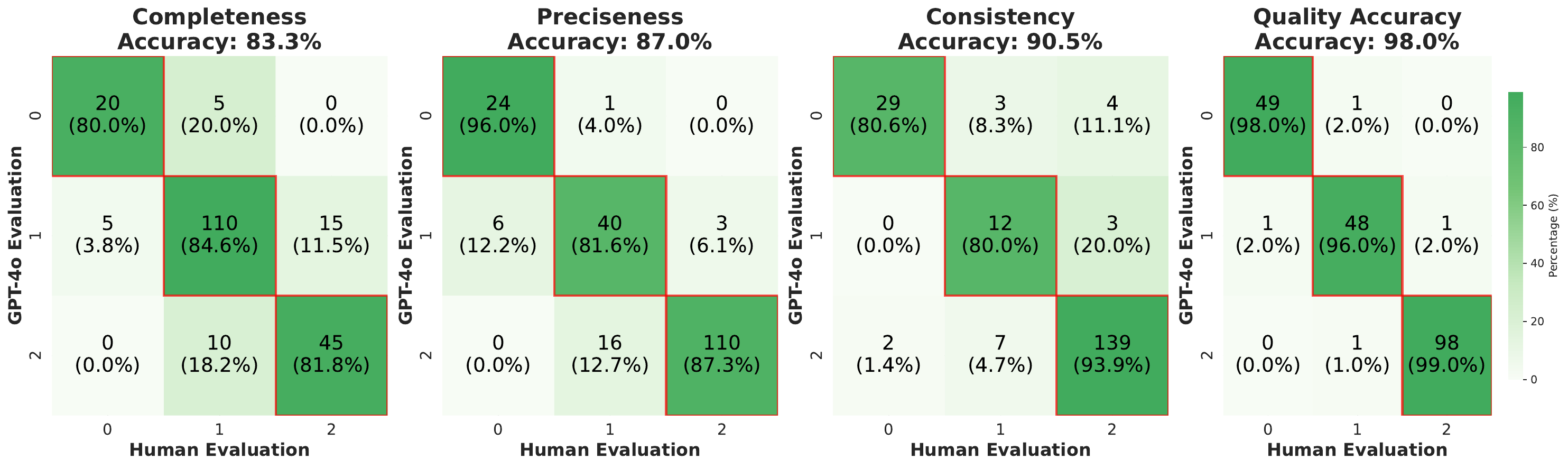}
    \caption{Confusion matrices showing alignment between human expert scores and GPT-4o automated evaluation across four evaluation dimensions.}
    \label{fig:confusion_matrices}
\end{figure*}

The confusion matrices shown in Figure~\ref{fig:confusion_matrices} demonstrate strong alignment between human expert scores and GPT-4o automated evaluation across all three evaluation dimensions, with over 80\% accuracy in each dimension. Quadratic weighted $\kappa_w$ accounts for the ordinal nature of the evaluation labels, penalizing larger discrepancies more heavily than adjacent category differences. The consistently high $\kappa_w$ values (0.774–0.985) detailed in Table~\ref{tab:kappa_icc_appendix} indicate substantial agreement beyond chance between human expert scores and GPT-4o automated evaluation, reflecting that the automated system is not only accurate but also aligned with the fine-grained ordinal structure of human expert judgments. 

\begin{table}[!ht]
\centering
\footnotesize
\resizebox{\linewidth}{!}{%
\begin{tabular}{lcccc}
\toprule
\textbf{Metric} & Completeness & Preciseness & Consistency & Quality Accuracy \\
\midrule
\textbf{$\kappa_w$} & 0.774 & 0.876 & 0.840 & 0.985 \\
\bottomrule
\end{tabular}}
\caption{Quadratic weighted Cohen's $\kappa_w$ values for human–AI alignment across evaluation dimensions.}
\label{tab:kappa_icc_appendix}
\end{table}

To assess whether human–AI alignment remains stable across quality levels, we partitioned the 200 sampled outputs into three tiers by human-score quantiles and examined the lowest third, which contains more low-quality or failed predictions. Table~\ref{tab:alignment_lowest_tier} presents the alignment metrics in this challenging subset. While the weighted $\kappa_w$ values show some decrease compared to overall alignment, the automated evaluation does not substantially degrade even under these challenging cases, with $\kappa_w$ values remaining above 0.7 for most dimensions and achieving 0.952 for Quality Accuracy. This confirms that our automated evaluation framework maintains robust alignment with human expert annotations, strengthening confidence in its use as a reliable surrogate for large-scale human evaluation.

\begin{table}[!htbp]
\centering
\small
\resizebox{\linewidth}{!}{%
\begin{tabular}{lcccc}
\toprule
\textbf{Metric} & Completeness & Preciseness & Consistency & Quality Accuracy \\
\midrule
\textbf{$\kappa_w$} & 0.702 & 0.791 & 0.755 & 0.952 \\
\bottomrule
\end{tabular}}
\caption{Human–AI alignment in the lowest-quality tier (bottom third by human score). Despite challenging cases, alignment remains robust across all evaluation dimensions.}
\label{tab:alignment_lowest_tier}
\end{table}

\begin{table*}[!tbp]
\centering
\small
\resizebox{\textwidth}{!}{
\begin{tabular}{lccccccc}
\toprule
\textbf{Model} & \textbf{GPT-4o} & \textbf{Claude} & \textbf{Gemini} & \textbf{ICC} & \textbf{ICC 95\% CI} & \textbf{MAD (Claude)} & \textbf{MAD (Gemini)} \\
\midrule
\rowcolor{green!8} Qwen2.5-VL-Instruct (7B) & 4.367 & 4.376 & 4.371 & 0.984 & [0.980, 0.987] & 0.009 & 0.004 \\
\rowcolor{green!8} Qwen2.5-VL-Instruct (32B) & 5.272 & 5.294 & 5.308 & 0.992 & [0.990, 0.994] & 0.022 & 0.036 \\
\rowcolor{green!8} InternVL3 (8B) & 4.983 & 4.995 & 4.968 & 0.986 & [0.982, 0.989] & 0.012 & 0.015 \\
\rowcolor{green!8} InternVL3 (38B) & 4.965 & 4.977 & 4.977 & 0.990 & [0.988, 0.992] & 0.012 & 0.012 \\
\rowcolor{green!8} Qwen2.5-VL-Instruct (72B) & 4.982 & 4.932 & 4.959 & 0.988 & [0.985, 0.990] & 0.050 & 0.023 \\
\midrule
\rowcolor{blue!8} BiMediX2 (8B) & 1.721 & 1.692 & 1.719 & 0.954 & [0.942, 0.963] & 0.029 & 0.002 \\
\rowcolor{blue!8} Lingshu (32B) & 4.312 & 4.344 & 4.317 & 0.987 & [0.984, 0.990] & 0.032 & 0.005 \\
\rowcolor{blue!8} MedGemma (27B) & 4.054 & 4.036 & 4.036 & 0.978 & [0.973, 0.983] & 0.018 & 0.018 \\
\midrule
\rowcolor{orange!15} Mistral-Medium-3 & 4.557 & 4.529 & 4.520 & 0.988 & [0.985, 0.990] & 0.028 & 0.037 \\
\rowcolor{orange!15} Claude-4-Sonnet & 4.529 & 4.529 & 4.570 & 0.981 & [0.976, 0.985] & 0.000 & 0.041 \\
\rowcolor{orange!15} Gemini-2.5-Pro & 5.018 & 5.014 & 5.005 & 0.985 & [0.981, 0.988] & 0.004 & 0.013 \\
\rowcolor{orange!15} GPT-4o & 5.321 & 5.335 & 5.321 & 0.989 & [0.986, 0.991] & 0.014 & 0.000 \\
\rowcolor{orange!15} GPT-5 & 5.679 & 5.679 & 5.652 & 0.987 & [0.983, 0.989] & 0.000 & 0.027 \\
\bottomrule
\end{tabular}
}
\caption{Multi-judge validation results showing overall description scores from three independent LLM judges and inter-judge consistency metrics. MAD (Claude) and MAD (Gemini) represent MAD between GPT-4o scores and Claude/Gemini scores respectively. Score differences across judges are minimal, and high ICC values ($>$0.95) confirm excellent agreement, indicating no systematic self-preference bias.}
\label{tab:multijudge_validation}
\end{table*}

\paragraph{Multi-Judge Validation for Self-Preference Analysis.}
To investigate potential self-preference bias when using GPT-4o as the evaluation judge, we conducted a multi-judge validation study using three independent LLM judges: GPT-4o, Claude-4-Sonnet, and Gemini-2.5-Pro. Each judge independently scored all model outputs on the No-Reference description tasks. Table~\ref{tab:multijudge_validation} presents the overall scores assigned by each judge and inter-judge consistency metrics. The results show minimal variation between judges, with score differences typically within 0.05 points. Notably, GPT-4o does not assign systematically higher scores to OpenAI models (GPT-4o, GPT-5) compared to scores from Claude or Gemini judges. All models achieve Intraclass Correlation Coefficient (ICC) values above 0.95, indicating excellent agreement among judges. The mean absolute deviation (MAD) between GPT-4o scores and other judges remains below 0.05 for most models, confirming that GPT-4o's evaluations are consistent with independent judges and do not exhibit self-preference bias.

\section{Potential Risks}

Although MedQ-Bench is intended for benchmark evaluation rather than direct clinical deployment, careful interpretation is still important. Because the benchmark is task-agnostic by design, strong performance should be viewed as evidence of foundational quality reasoning, not as direct proof of task-specific clinical readiness. In addition, current MLLMs may show variable performance on subtle degradations, fine-grained comparisons, and complex quality descriptions, and this variability should be considered when generalizing results. We also observe that some model families emphasize different strengths, for example stronger perception with relatively weaker description in certain settings. Overall, benchmark outcomes are most informative when combined with task-aware validation and expert review in downstream scenarios.

\section{Artifact Licensing and Intended Use}
\label{sec:appendix_licensing}

All public datasets reused in \medqbench{} (Radiopaedia, AAPM CT-MAR~\citep{haneda2025aapm}, MR-ART~\citep{narai2022movement}, FSL Example MRI Artifacts, FastMRI~\citep{zbontar2018fastmri}, HistoArtifacts and HARP~\citep{fuchs2024harp}, EndoCV2020~\citep{polat2020endoscopic}, EyeQ~\citep{fu2019evaluation}) are accessed under their original licenses (research-only or CC-BY variants) and used strictly within those terms. The private 5T MRI subset was collected under institutional ethical approval with full patient anonymization. All evaluated MLLMs are queried via their official APIs or Hugging Face checkpoints in accordance with each provider's terms of service. \medqbench{} itself, including the curated images, generated questions, and expert annotations, is released for non-commercial academic research only, inheriting the most restrictive upstream license; derivatives may not be used in clinical deployment or commercial products. This usage is consistent with the intended research purpose of every upstream artifact, namely advancing medical image quality assessment and the evaluation of MLLMs.

\clearpage
\section{Evaluation Prompt}

\begin{tcolorbox}[breakable,colback=gray!5,colframe=black!60,title=Single-Image Perception Task Prompts,fontupper=\scriptsize,fonttitle=\footnotesize\bfseries]

\textbf{Yes-No / What / How Question Template:}
\begin{tcolorbox}[colback=gray!10,colframe=gray!50,boxrule=0.5pt,arc=2pt,fontupper=\scriptsize]
\ttfamily
You are an expert in medical image quality assessment. Please carefully observe this medical image and answer the following question:
\end{tcolorbox}
\end{tcolorbox}

\begin{tcolorbox}[breakable,colback=blue!5,colframe=blue!60,title=Description Task Prompts,fontupper=\scriptsize,fonttitle=\footnotesize\bfseries]

\textbf{No-reference Description Template:}
\begin{tcolorbox}[colback=gray!10,colframe=gray!50,boxrule=0.5pt,arc=2pt,fontupper=\scriptsize]
\ttfamily
As a medical image quality assessment expert, provide a concise description focusing on low-level appearance of the image in details. Conclude with ``Overall, the quality of this image is [good/usable/reject]''. Please provide a comprehensive but concise assessment in 3-5 sentences.
\end{tcolorbox}

\textbf{Comprehensive Description Template:}
\begin{tcolorbox}[colback=gray!10,colframe=gray!50,boxrule=0.5pt,arc=2pt,fontupper=\scriptsize]
\ttfamily
As a medical image quality assessment expert, provide a concise description comparing two images focusing on low-level appearance. Conclude with which image has higher quality. Please provide comprehensive but concise assessment in 3-5 sentences.
\end{tcolorbox}
\end{tcolorbox}

\begin{tcolorbox}[colback=gray!5,colframe=black!60,title=Complete Evaluation Prompt Templates for No-reference Description Tasks,fontupper=\footnotesize,fonttitle=\small\bfseries,breakable]

\textbf{Completeness Evaluation Prompt:}
\begin{Verbatim}[breaklines=true,breakanywhere=true,fontsize=\scriptsize]
#System: You are a helpful assistant.
#User: Evaluate whether the description [MLLM DESC] completely includes the low-level visual information in the reference description [GOLDEN DESC]. Please rate score 2 for completely or almost completely including reference information, 0 for not including at all, 1 for including part of the information or similar description. Please only provide the result in the following format: Score:
\end{Verbatim}

\textbf{Preciseness Evaluation Prompt:}
\begin{Verbatim}[breaklines=true,breakanywhere=true,fontsize=\scriptsize]
#System: You are a helpful assistant.
#User: The precision metric evaluates whether the low-level description is consistent with the reference and reasonably aligned with the final quality judgment. Minor wording differences or small omissions that do not change the overall meaning should still be considered consistent. Only penalize clear contradictions with the reference, such as describing blur for clear, noisy for clean, motion-free for motion artifacts, noise-free for low-dose noise, etc. Evaluate whether output [MLLM DESC] reasonably reflects reference [GOLDEN DESC]. Please rate score 2 for overall consistency and no major contradictions with the quality conclusion, 1 for partial consistency or very few minor contradictions, and 0 for obvious contradictions or misalignment with the quality conclusion. Please only provide the result in the following format: Score:
\end{Verbatim}

\textbf{Consistency Evaluation Prompt:}
\begin{Verbatim}[breaklines=true,breakanywhere=true,fontsize=\scriptsize]
#System: You are a helpful assistant.
#User: Evaluate the internal consistency between the description path (description of image problems) and the final quality judgment in [MLLM DESC]. The description should logically support the final quality conclusion. For example, if many serious problems are described, the final quality should be "reject"; if minor problems are described, it should be "usable"; if no or very few problems are described, it should be "good". Compare with the reference [GOLDEN DESC] to understand the expected description-conclusion relationship. Please rate score 2 for highly consistent description and conclusion, 1 for partially consistent with minor logical gaps, and 0 for major inconsistency between described problems and quality judgment. Please only provide the result in the following format: Score:
\end{Verbatim}

\textbf{Quality Accuracy Evaluation Prompt:}
\begin{Verbatim}[breaklines=true,breakanywhere=true,fontsize=\scriptsize]
#System: You are a helpful assistant.
#User: Evaluate the accuracy of the final quality judgment in [MLLM DESC] compared to the reference [GOLDEN DESC]. The quality levels have a progressive relationship: reject < usable < good. Consider the distance between predicted and reference quality: Please rate score 2 for exactly matching the reference quality level, 1 for adjacent level difference (e.g., usable vs good, or reject vs usable), and 0 for distant level difference (reject vs good) or completely incorrect quality assessment. Please only provide the result in the following format: Score:
\end{Verbatim}
\end{tcolorbox}

\begin{tcolorbox}[colback=gray!5,colframe=black!60,title=Complete Evaluation Prompt Templates for Comparison Description Tasks,fontupper=\footnotesize,fonttitle=\small\bfseries,breakable]

\textbf{Completeness Evaluation Prompt:}
\begin{Verbatim}[breaklines=true,breakanywhere=true,fontsize=\scriptsize]
#System: You are a helpful assistant.
#User: Evaluate whether the description [MLLM DESC] completely includes the low-level visual information in the reference description [GOLDEN DESC]. Please rate score 2 for completely or almost completely including reference information, 0 for not including at all, 1 for including part of the information or similar description. Please only provide the result in the following format: Score:
\end{Verbatim}

\textbf{Preciseness Evaluation Prompt:}
\begin{Verbatim}[breaklines=true,breakanywhere=true,fontsize=\scriptsize]
#System: You are a helpful assistant.
#User: The precision metric evaluates whether the low-level description is consistent with the reference and reasonably aligned with the final quality judgment. Minor wording differences or small omissions that do not change the overall meaning should still be considered consistent. Only penalize clear contradictions with the reference, such as describing blur for clear, noisy for clean, motion-free for motion artifacts, noise-free for low-dose noise, etc. Evaluate whether output [MLLM DESC] reasonably reflects reference [GOLDEN DESC]. Please rate score 2 for overall consistency and no major contradictions with the quality conclusion, 1 for partial consistency or very few minor contradictions, and 0 for obvious contradictions or misalignment with the quality conclusion. Please only provide the result in the following format: Score:
\end{Verbatim}

\textbf{Consistency Evaluation Prompt:}
\begin{Verbatim}[breaklines=true,breakanywhere=true,fontsize=\scriptsize]
#System: You are a helpful assistant.
#User: Evaluate the internal consistency between the description path (comparative description of image problems) and the final quality comparison judgment in [MLLM DESC]. The description should logically support the final comparison conclusion. Compare with the reference [GOLDEN DESC] to understand the expected description-conclusion relationship for image comparison. Please rate score 2 for highly consistent description and comparison conclusion, 1 for partially consistent with minor logical gaps, and 0 for major inconsistency between described comparative problems and quality comparison judgment. Please only provide the result in the following format: Score:
\end{Verbatim}

\textbf{Quality Accuracy Evaluation Prompt:}
\begin{Verbatim}[breaklines=true,breakanywhere=true,fontsize=\scriptsize]
#System: You are a helpful assistant.
#User: Evaluate the accuracy of the final quality comparison judgment in [MLLM DESC] compared to the reference [GOLDEN DESC]. The comparison should correctly identify which image has higher quality based on the described visual characteristics. Please rate score 2 for exactly matching the reference quality comparison, and 0 for completely incorrect quality comparison (opposite conclusion) or unreasonable assessment. Please only provide the result in the following format: Score:
\end{Verbatim}

\end{tcolorbox}

\end{document}